\documentclass[10pt,twocolumn,letterpaper]{article}

\usepackage[pagenumbers]{cvpr}              %

\usepackage{url}            %
\usepackage{booktabs}       %
\usepackage{amsfonts}       %
\usepackage{nicefrac}       %
\usepackage{microtype}      %
\usepackage{xcolor}         %
\usepackage{graphicx}
\usepackage{amsmath}
\usepackage{amssymb}
\usepackage{multirow}
\usepackage{threeparttable}
\usepackage{enumitem}
\usepackage{float}
\usepackage{caption}
\usepackage{subcaption}
\usepackage{algorithm}
\usepackage{algorithmic}
\usepackage{bm}
\usepackage{mathbbol}
\usepackage{amsmath, amssymb, bm}
\usepackage{wrapfig}
\usepackage{lipsum}
\usepackage{tikz}
\usetikzlibrary{positioning}
\usetikzlibrary{shapes,arrows}
\usetikzlibrary{calc}
\usetikzlibrary{decorations.pathreplacing}
\usetikzlibrary{angles,quotes}
\usetikzlibrary{patterns}
\usetikzlibrary{3d}
\usepackage{circuitikz}
\usepackage[percent]{overpic}
\usepackage{tikz,pgfplots}
\pgfplotsset{compat=1.18}
\definecolor{barPink}{HTML}{D0A215}
\definecolor{barOrange}{HTML}{BC5215}
\usetikzlibrary{calc}
\definecolor{myGreen}{HTML}{879A39}
\usepackage{indentfirst}

\usepackage[accsupp]{axessibility}  %

\definecolor{cvprblue}{rgb}{0.21,0.49,0.74}
\usepackage[pagebackref,breaklinks,colorlinks,allcolors=cvprblue]{hyperref}

\def\paperID{13755} %
\def\confName{CVPR}
\def\confYear{2026}

\title{\METHOD: Efficient Large-Scale Physical Modeling \\ via Parallelized Multi-Scale Attention}

\author{Pedro M. P. Curvo \quad Jan-Willem van de Meent \quad Maksim Zhdanov\\
University of Amsterdam\\
{\tt\small pedro.pombeiro.curvo@student.uva.nl}
}

\newcommand{\METHOD}{\textsc{MSPT}}

\begin{document}
\maketitle
\begin{abstract}
A key scalability challenge in neural solvers for industrial-scale physics simulations is efficiently capturing both fine-grained local interactions and long-range global dependencies across millions of spatial elements. We introduce the Multi-Scale Patch Transformer (MSPT)\footnote{\href{https://github.com/pedrocurvo/mspt}{https://github.com/pedrocurvo/mspt}}, an architecture that combines local point attention within patches with global attention to coarse patch-level representations. To partition the input domain into spatially-coherent patches, we employ ball trees, which handle irregular geometries efficiently. This dual-scale design enables MSPT to scale to millions of points on a single GPU. We validate our method on standard PDE benchmarks (elasticity, plasticity, fluid dynamics, porous flow) and large-scale aerodynamic datasets (ShapeNet-Car, Ahmed-ML), achieving state-of-the-art accuracy with substantially lower memory footprint and computational cost.

\end{abstract}
    
\vspace{-1em}
\section{Introduction}
\label{sec:intro}

Deep learning is increasingly applied to model complex physical phenomena as a computationally efficient surrogate. However, as applications in computational fluid dynamics (CFD) and multi-physics design scale to simulations with millions of mesh points, these methods face substantial computational challenges~\cite{alkin2025abuptscalingneuralcfd}. Developing neural architectures capable of efficiently processing millions of spatial elements has therefore become important~\cite{alkin2025abuptscalingneuralcfd}, particularly for high-throughput environments such as large-scale design optimization and real-time industrial analysis where both accuracy and scalability are required. These challenges require neural surrogates that capture fine-grained physical interactions while maintaining computational efficiency. %

\newsavebox{\roundedimg}
\sbox{\roundedimg}{%
  \includegraphics[width=\linewidth,trim=0 0mm 0 28mm]{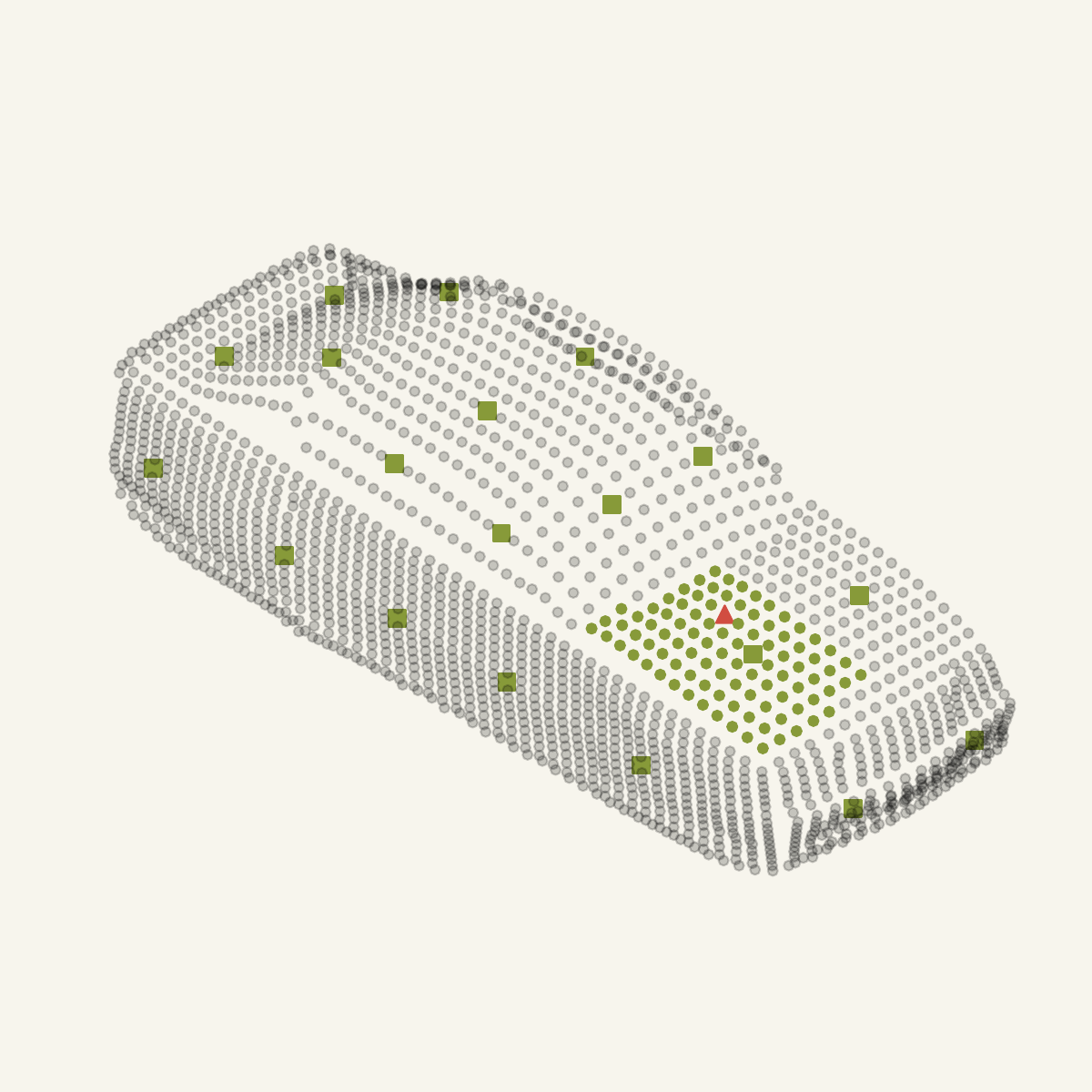}%
}

\begin{figure}
  \centering
  \begin{tikzpicture}
    \clip[rounded corners=10pt]
      (0,0) rectangle (\wd\roundedimg,\ht\roundedimg+\dp\roundedimg);
    \node[anchor=south west, inner sep=0pt] at (0,0) {\usebox{\roundedimg}};
    \draw[line width=1pt, rounded corners=10pt, black]
      (0,0) rectangle (\wd\roundedimg,\ht\roundedimg+\dp\roundedimg);

    \pgfmathsetmacro{\xoffset}{0.10*\wd\roundedimg/1pt}
    \pgfmathsetmacro{\yoffset}{0.15*\ht\roundedimg/1pt}
    \pgfmathsetmacro{\plotwidth}{0.35*\wd\roundedimg/1pt}
    \pgfmathsetmacro{\plotheight}{0.25*\ht\roundedimg/1pt}

    \begin{axis}[
        at={(\xoffset-7 pt,\yoffset-4 pt)}, %
        anchor=south west,
        width=\plotwidth pt,
        height=\plotheight pt,
        xbar,
        symbolic y coords={Patch A, Patch B},
        ytick=\empty,
        xtick=\empty,
        axis line style={draw=none},
        tick style={draw=none},
        bar width=8pt,
        y=0.5cm,
        enlarge y limits=0.3,
        clip=false,
        opacity=0.95,
        xmin=0, xmax=3,
    ]
        \addplot[fill=barPink]  coordinates {(2.8,Patch A)};
        \addplot[fill=barOrange]  coordinates {(4.1,Patch A)};
        \addplot[fill=barPink] coordinates {(2.36,Patch B)};
        \addplot[fill=barOrange] coordinates {(5.98,Patch B)};
    \end{axis}

    \filldraw[fill=myGreen, draw=black, line width=0.3pt]
  ($(7.05cm,5.95cm) + (-0.05cm,-0.05cm)$)
  rectangle ++(0.2cm,0.2cm);

    \node[anchor=south, font=\small, text=black] at (1.82cm,2.20cm) {Peak Memory (GB)};
    \node[anchor=south, font=\small, text=black] at (3.60cm,1.87cm) {42.8};
    \node[anchor=south, font=\small, text=black] at (1.98cm,1.50cm) {26.0};
    \node[anchor=south, font=\small, text=black] at (1.40cm,1.0cm) {Latency (ms)};
    \node[anchor=south, font=\small, text=black] at (2.65cm,0.67cm) {31};
    \node[anchor=south, font=\small, text=black] at (2.05cm,0.37cm) {28};

    \draw[
      dashed,  %
      line width=0.8pt,
      color=black!80
    ] (5.45cm,4.05cm) -- (7.1cm,5.9cm);  %
    \draw[
      dashed,  %
      line width=0.8pt,
      color=black!80
    ] (5.82cm,2.62cm) -- (7.1cm,5.9cm);  %
    \draw[
      dashed,  %
      line width=0.8pt,
      color=black!80
    ] (6.62cm,3.12cm) -- (7.1cm,5.9cm);  %
    \draw[
      dashed,  %
      line width=0.8pt,
      color=black!80
    ] (4.52cm,3.62cm) -- (7.1cm,5.9cm);  %
    \node[anchor=south, font=\small, text=black] at (6.2cm,5.5cm) {Pool};

    \draw[rounded corners=3pt, draw=black, line width=0.3pt]
      ($(5.9cm,0.55cm) + (-0.1cm,-0.2cm)$)  %
      rectangle ($(7.70cm,1.15cm) + (0.15cm,0.1cm)$);  %
        
    \filldraw[fill=barOrange, draw=black, line width=0.3pt]
      ($(6.05cm,0.95cm) + (-0.05cm,-0.05cm)$)  %
      rectangle ++(0.2cm,0.2cm);
    \node[anchor=south, font=\small, text=black] at (6.95cm,0.77cm) {Transolver};  %
    
    \filldraw[fill=barPink, draw=black, line width=0.3pt]
      ($(6.05cm,0.55cm) + (-0.05cm,-0.05cm)$)  %
      rectangle ++(0.2cm,0.2cm);
    \node[anchor=south, font=\small, text=black] at (6.73cm,0.38cm) {MSPT};  %
    
  \end{tikzpicture}

  \caption{Parallelized Multi-Scale Attention mechanism. Each patch performs local self-attention, while pooled supernodes exchange information globally across patches in parallel. Peak memory (GB) and latency (ms) on 500k points with 256 slices (\textit{Transolver}) and 256 patches (\textit{MSPT}).}
  \label{fig:high_level_view}
\end{figure}
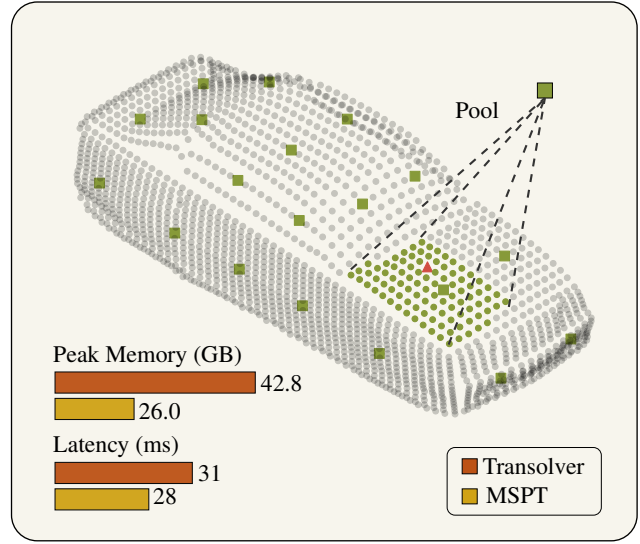
A key obstacle in scaling these methods lies in capturing spatial dependencies 
appropriate to different physical regimes. In solid mechanics, stress and strain fields localize around applied loads. By contrast, in incompressible fluids a global pressure coupling arises from the divergence-free constraint. Similarly, aerodynamics problems require computing forces from surface pressures, which must be consistent with far-field boundary conditions. Modeling such long-range dependencies poses a challenge, since computing all pairwise interactions scales quadratically with the system size.

Modeling long-range 
dependencies at industrial scale therefore requires approximations that preserve 
accuracy while reducing computational cost ~\citep{Majumdar2021}.
Recent approaches have made substantial progress toward scalable neural PDE solvers. Neural operator models~\citep{FNO, anonymous2023factorized, Gupta2021MultiwaveletbasedOL} have demonstrated that learning mappings between function spaces enables mesh-independent surrogate models. However, these spectral methods require structured grids or periodic boundary conditions and often struggle to accurately represent sharp, localized features without dense sampling or large frequency bases~\cite{wu2024Transolver}.

Transformer-based solvers~\cite{Cao2021ChooseAT,hao2023gnot,li2023transformer,anonymous2023improved} extend operator-learning concepts to unstructured data, including meshes and point clouds, by leveraging attention to capture long-range dependencies. A key area of innovation are architectures that avoid quadratic complexity. Transolver~\cite{wu2024Transolver} projects the input mesh to a reduced set of learnable global representations (slices), achieving sub-linear complexity. However, this approach scales poorly with bottleneck size and relies on pooling that compromises simulation fidelity.
UPT~\cite{alkin2024upt} introduces a centralized latent space that summarizes the domain via cross-attention between mesh points and shared latent tokens, providing global context applied uniformly across regions. Extending this idea, AB-UPT~\cite{alkin2025abuptscalingneuralcfd} adopts a branched design for CFD, splitting computation into surface and volume branches that perform self-attention independently and exchange information through cross-attention.
Erwin~\cite{zhdanov2025erwin} removes pooling entirely and instead achieves linear complexity by localizing attention to ball-tree partitions, preserving local fidelity. However, this approach suffers from limited information propagation in capturing long-range dependencies.

In this work, we introduce the Multi-Scale Patch Transformer (\METHOD), a parallelized multi-scale attention architecture that bridges fine-grained local interactions with global context in a scalable manner. \METHOD\ first partitions the domain into local patches of points, where attention is computed efficiently to capture fine-scale interactions within each region. To capture long-range interactions, we compute a coarse-grained representation of each patch and apply fine-to-coarse attention, allowing information to flow globally without quadratic cost. This multi-scale organization enables \METHOD\ to capture local and global dependencies within a single, unified attention operation, achieving near-linear scaling with the number of points.

The main contributions of this work are:
\begin{itemize}[leftmargin=15pt, topsep=0pt, itemsep=4pt]
\item We propose the Parallelized Multi-Scale Attention (PMSA) mechanism, which processes local patch-level interactions and global cross-patch interactions in parallel within a unified attention operation, enabling scalable operator learning with near-linear complexity.
\item We introduce \METHOD, a multi-block transformer architecture that handles arbitrary geometries and varying resolutions through flexible domain partitioning and hierarchical pooling.
\item We demonstrate state-of-the-art accuracy on standard PDE benchmarks (elasticity, plasticity, fluid dynamics, porous media flow) and industry-scale 3D aerodynamic datasets (ShapeNet-Car, Ahmed-ML) at significantly lower computational cost, potentially scaling to millions of points on a single GPU.
\end{itemize}

\section{Background}
\label{sec:background}

\paragraph{Neural Operator Learning}
Neural operators~\cite{Kovachki2023NeuralOL} are a class of deep learning models designed to learn mappings between infinite-dimensional function spaces, making them particularly well-suited for parameterizing neural PDE solvers~\cite{FNO}.
The key idea is to formulate a model as a sequence of linear integral operators computed in the spectral (Fourier) domain, thus guaranteeing discretization invariance. The main limitations of the original work - specifically suboptimal local feature representations and reliance on uniform grids - were addressed in subsequent works~\cite{rahman2022u, Guibas2021AdaptiveFN, li2023geometryinformed}.
In general, spectral methods efficiently capture global dependencies through frequency-domain operations and achieve strong performance on regular, periodic domains.

\paragraph{Transformer-based PDE Solvers}
Complementing spectral approaches, transformer-based architectures leverage attention mechanisms to handle unstructured data such as meshes and point clouds~\citep{Cao2021ChooseAT, liu2022htnet, hao2023gnot, li2023scalable, li2023transformer, anonymous2023improved, holzschuh2025pdetransformerefficientversatiletransformers}. Self-attention naturally models long-range dependencies across irregular domains without requiring regular grid structure. For example, GNOT~\citep{hao2023gnot} augments attention with topology-aware neighborhoods, while OFormer~\citep{li2023transformer} formulates PDE solution inference as a sequence-to-sequence problem using cross-attention between spatial queries and target fields. CViT~\cite{Wang2024CViTCV} leverages recent advancements in computer vision and builds upon a ViT~\cite{vit} encoder to capture multi-scale dependencies. Overall, transformer-based approaches achieve state-of-the-art performance on most PDE-related tasks.

\paragraph{Hierarchical Models}
To address the scalability bottleneck of attention, hierarchical formulations introduce spatial structure and multi-scale reasoning. ViT~\cite{vit} first demonstrated the effectiveness of patch-based tokenization in vision, while PatchFormer~\cite{patchformer} extended this idea to point clouds by clustering points into compact patch tokens that approximate global interactions, and subsequent work further emphasized preserving global-local interactions via point cloud tokenization~\cite{KHAN2024110712}. For physical systems, HT-Net~\cite{liu2022htnet} constructs explicit coarse-to-fine representations to capture dynamics at multiple scales,and domain-decomposition approaches in physics learning (e.g., PhyFlow~\cite{9679054} and XPINNs~\cite{jagtap2020extended}) similarly combine local modeling with global consistency. Likewise, UPT~\cite{alkin2024upt} introduces a latent operator framework where learnable tokens summarize the global domain through cross-attention with physical points, providing compact global context.
Erwin~\cite{zhdanov2025erwin} imposes regular structure on irregular point cloud data using ball trees, which enables sparse attention and achieves linear complexity.
Building upon UPT~\cite{alkin2024upt}, AB-UPT~\cite{alkin2025abuptscalingneuralcfd} introduces a two-branch transformer architecture specifically designed for CFD domains. The model splits computation into a \emph{surface branch} and a \emph{volume branch}, allowing each to specialize in distinct physical regimes. Each branch processes its subset of mesh points with self-attention, while cross-attention between branches enables bidirectional information exchange.

\section{Multi-Scale Patch Transformer (\METHOD)}
\label{sec:mspt}
We present our Parallelized Multi-Scale Attention mechanism and the overall \METHOD\ architecture built upon it.

\begin{mdframed}[
  backgroundcolor=blue!5,
  linecolor=blue!40,
  linewidth=1pt,
  roundcorner=5pt,
  innertopmargin=8pt,
  innerbottommargin=8pt,
  innerleftmargin=8pt,
  innerrightmargin=8pt
]
\textbf{Core idea.} 
We partition a point cloud into patches. Points attend locally within their patch and globally to pooled patch representations, capturing both fine-grained structure and long-range context.
\end{mdframed}

\subsection{Parallelized Multi-Scale Attention Mechanism}
\label{sec:patched-attention}

We begin with a point cloud and a feature matrix
\[
\mathcal{P} = \{\mathbf{p}_1, \ldots, \mathbf{p}_N\} \subset \mathbb{R}^D, \quad \mathbf{H} \in \mathbb{R}^{N \times F},
\]
where $F$ is the feature dimension. We partition the point cloud into $K$ non-overlapping patches of size $L$,\footnote{Padding is applied as needed so that $N$ is divisible by $L$.} so that
\[
\mathcal{P} = \bigcup_{k=1}^{K} \mathcal{P}_k, \qquad |\mathcal{P}_k| = L, \qquad N = K L,
\]
with feature blocks $\mathbf{H}_k \in \mathbb{R}^{L \times F}$ for each patch.

In the case of unstructured grids, we employ Ball Tree partitioning \cite{zhdanov2025erwin}: by building a balanced ball tree over $\mathcal{P}$ and reordering points according to a depth-first traversal of its leaves, we obtain contiguous blocks of length $L$ as patches $\mathcal{P}_k$ (see Section~\ref{sec:patchouli-architecture} and Appendix~\ref{app:balltree} for details).

\paragraph{Pooled global context.}
In each patch $\mathcal{P}_k$, we summarize the $L$ local tokens into $Q$ pooled tokens using a pooling operator $\mathrm{Pool}:\mathbb{R}^{L \times F}\rightarrow\mathbb{R}^{Q \times F}$ (see Section~\ref{sec:patchouli-architecture}):
\begin{equation}
\mathbf{S}_k = \mathrm{Pool}(\mathbf{H}_k) \in \mathbb{R}^{Q \times F}.
\end{equation}
Stacking all patch summaries row-wise yields the global context matrix: %
\begin{equation}
\mathbf{S} = \begin{bmatrix} \mathbf{S}_1; & \mathbf{S}_2; & \cdots; & \mathbf{S}_K \end{bmatrix} \in \mathbb{R}^{(KQ) \times F}.
\end{equation}
In practice, we can adjust the patch size $L$ such that $KQ~\ll~N$. This allows for efficient inter-patch communication without substantial computational overhead, as we will explain in the following section.

\paragraph{Augmented dual-scale attention.}
We then augment local block features $\mathbf{H}_k$ with the   context matrix:
\begin{equation}
\mathbf{Z}_k = \begin{bmatrix} \mathbf{H}_k \\ \mathbf{S} \end{bmatrix} \in \mathbb{R}^{(L + KQ) \times F}.
\end{equation}

This formulation allows us to compute local interactions as well as capture global dependencies within a single self attention \cite{Vaswani2017AttentionIA} operation:
\begin{gather}
\mathbf{Q}_k = \mathbf{Z}_k \mathbf{W}_Q, \qquad
\mathbf{K}_k = \mathbf{Z}_k \mathbf{W}_K, \qquad
\mathbf{V}_k = \mathbf{Z}_k \mathbf{W}_V, \notag
\\
\mathbf{A}_k = \mathrm{softmax}\!\left(\frac{\mathbf{Q}_k \mathbf{K}_k^\top}{\sqrt{F}}\right)
\end{gather}

To expose the dual-scale structure, we partition the attention and value matrices according to the block structure of $\mathbf{Z}_k$. The attention matrix decomposes as:
\begin{equation}
\mathbf{A}_k = \begin{bmatrix} 
\mathbf{A}_k^{\text{loc,loc}} & \mathbf{A}_k^{\text{loc,glob}} \\
\mathbf{A}_k^{\text{glob,loc}} & \mathbf{A}_k^{\text{glob,glob}}
\end{bmatrix},
\end{equation}
where $\mathbf{A}_k^{\text{loc,loc}} \in \mathbb{R}^{L \times L}$ captures local-to-local attention within the patch, $\mathbf{A}_k^{\text{loc,glob}} \in \mathbb{R}^{L \times KQ}$ captures local-to-global attention from patch tokens to all pooled tokens, and the remaining blocks $\mathbf{A}_k^{\text{glob,loc}} \in \mathbb{R}^{KQ \times L}$ and $\mathbf{A}_k^{\text{glob,glob}} \in \mathbb{R}^{KQ \times KQ}$ govern how global tokens attend to local and global features, respectively.

Similarly, we partition the value matrix:
\begin{equation}
\mathbf{V}_k = \begin{bmatrix} \mathbf{V}_k^{\text{loc}} \\ \mathbf{V}_k^{\text{glob}} \end{bmatrix}, \quad 
\mathbf{V}_k^{\text{loc}} \in \mathbb{R}^{L \times F}, \quad
\mathbf{V}_k^{\text{glob}} \in \mathbb{R}^{KQ \times F}.
\end{equation}

The attention operation produces updated representations for both the local patch tokens and the global supernodes; we retain the former resulting in the following update of a token $i$ in block $k$:
\[
\mathbf{h}'_{k,i}
= \underbrace{\sum_{j=1}^{L} a^{\text{loc,loc}}_{k,ij}\,\mathbf{v}^{\text{loc}}_{k,j}}_{\substack{\text{local attention} \\ \text{within patch}}}
+
\underbrace{\sum_{m=1}^{KQ} a^{\text{loc,glob}}_{k,im}\,\mathbf{v}^{\text{glob}}_{m}}_{\substack{\text{global attention} \\ \text{via pooled tokens}}},
\]
or, in the matrix form,
\begin{equation}
\mathbf{H}_k' = \mathbf{A}_k^{\text{loc,loc}}\,\mathbf{V}_k^{\text{loc}}
+ \mathbf{A}_k^{\text{loc,glob}}\,\mathbf{V}_k^{\text{glob}}
\in \mathbb{R}^{L\times F}.
\end{equation}
The first term captures fine-grained structure within patch $k$, while the second provides long-range context via the pooled representations. After processing all patches, we stack the results to recover the full sequence:
\[
\mathbf{H}' = \begin{bmatrix} \mathbf{H}_1'; & \cdots; & \mathbf{H}_K' \end{bmatrix} %
\]

\paragraph{Multi-head extension.}
We extend to multi-head attention in the standard way \cite{Vaswani2017AttentionIA}. With $H_a$ attention heads, each head $h$ operates on a $(F/H_a)$-dimensional subspace with its own weight matrices $\mathbf{W}_Q^{(h)}, \mathbf{W}_K^{(h)}, \mathbf{W}_V^{(h)}$. The outputs are concatenated and projected:
\begin{equation}
\mathrm{MHA}(\mathbf{Z}_k) = \big[\mathbf{O}_k^{(1)} \,\|\, \cdots \,\|\, \mathbf{O}_k^{(H_a)}\big] \mathbf{W}_O.
\end{equation}
Retaining the first $L$ rows yields the updated patch features $\mathbf{H}_k'$, and stacking across patches recovers $\mathbf{H}'$.

\paragraph{Compact operator form.}
We summarize the full Parallelized Multi-Scale Attention (PMSA) mechanism as:
\begin{equation}
\boxed{
\mathbf{H}' = \mathrm{PMSA}(\mathbf{H})
     = \bigoplus_{k=1}^{K} \Pi_{\text{loc}} \, \mathrm{MHA}\left(\begin{bmatrix} \mathbf{H}_k \\ \mathbf{S} \end{bmatrix}\right).
}
\end{equation}
The operator $\Pi_{\text{loc}}$ retains the first $L$ rows (local tokens), and $\bigoplus_{k=1}^K$ stacks results across all patches.

\paragraph{Computational cost.} 
PMSA has complexity $O(NL + N^2Q/L)$, decomposed into local attention within patches and global attention to pooled tokens. While quadratic in $N$, the coefficient $Q/L$ of the quadratic term is typically small, making the linear term dominant in practice. The patch size $L$ offers a tunable trade-off: larger $L$ suppresses the quadratic global communication cost but increases local attention cost, while smaller $L$ reduces local computation at the expense of more cross-patch overhead. This flexibility enables PMSA to scale to point clouds with millions of points on a single GPU, see Section~\ref{sec:efficiency_analysis}.

\subsection{Model Architecture}
\label{sec:patchouli-architecture}
Building on Parallelized Multi-Scale Attention, we propose the Multi-Scale Patch Transformer (\METHOD).

\paragraph{Input Embedding and Preprocessing}
We start from a point cloud $\mathcal{P} = \{\mathbf{p}_1, \ldots, \mathbf{p}_N\} \subset \mathbb{R}^D$, where each point $\mathbf{p}_i$ has coordinates $\mathbf{x}_i$. For each point, we construct an input feature vector $\mathbf{r}_i$ by concatenating $\mathbf{x}_i$ with geometric descriptors $\mathbf{g}_i$. A shared embedding MLP $\phi_{\text{emb}}$ maps these raw features to hidden features $\mathbf{h}_i = \phi_{\text{emb}}(\mathbf{r}_i) \in \mathbb{R}^F$. We then pad the point set so that $N$ is divisible by the number of patches $K$, assigning zero features to padded entries.

\paragraph{Ball Tree partitioning}
We impose spatial locality on point clouds or unstructured meshes by building a balanced ball tree on the coordinates using the \texttt{balltree-erwin} implementation~\citep{zhdanov2025erwin}. The leaf order induces a spatially local permutation of points, and the patches are formed as contiguous blocks of length $L$ in this permuted sequence. This partitioning is computed once before the first transformer block and reused across blocks (see Appendix~\ref{app:balltree}).

\paragraph{Supernode Pooling}
Within each patch $k$, \METHOD\ compresses the $L$ point tokens into $Q$ supernode tokens $\mathbf{S}_k\in\mathbb{R}^{Q \times C}$ using one of the pooling operators discussed below. Let $q$ indicate a sub-patch $\mathbf{H}^q_k$  of size $L / Q$. We use mean or max pooling to aggregate information from each sub-patch to a corresponding supernode:
\vspace{0.2em}
\begin{equation}
\mathbf{S}^q_{k} = \frac{1}{L/Q} \sum_{j=1}^{L/Q} (\mathbf{H}^q_k)_{j}
\end{equation}
\begin{equation}
\mathbf{S}^q_{k} = \max_{j=1}^{L/Q} (\mathbf{H}^q_k)_{j}
\end{equation}
Alternatively, we employ linear projection:
\[
\mathbf{S}_k = \mathbf{W}_{\text{pool}}^\top \mathbf{H}_k,
\quad
\mathbf{W}_{\text{pool}} \in \mathbb{R}^{L \times Q},
\]
which learns a fixed set of $Q$ linear combinations of the $L$~tokens from the patch $k$.
The $K$ sets of $Q$ supernodes are concatenated to form the global supernode set $S \in \mathbb{R}^{KQ \times C}$.

\begin{figure}[t]
    \centering
    \begin{overpic}[width=\linewidth, trim=1.5em 1.5em 1.5em 1.5em, clip]{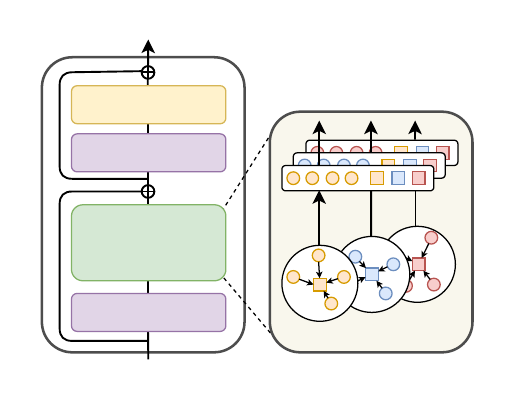}
    \put(22,55.5){\color{black}\small MLP}
    \put(17,45){\color{black}\small Layer Norm}
    \put(17,30){\color{black} \small Parallelized}
    \put(17,25.5){\color{black} \small Multi-Scale}
    \put(18,21){\color{black} \small Attention}
    \put(17,10){\color{black} \small Layer Norm}
    
    \put(57,31.5){\tikz{\node[fill=white, fill opacity=0.9, line width=0pt, inner sep=1pt, text opacity=1] {\small Multi-head Attention};}}

    \put(59,5.){\color{black}\small Ball Tree partitioning}
    \end{overpic}
    \caption{MSPT-Block. Each block partitions the point set into patches and pools local information into a small set of supernodes (here, 1). Multi-head attention is applied within each patch, augmented by the shared supernodes (global context).}
    \label{fig:architecture}
\end{figure}
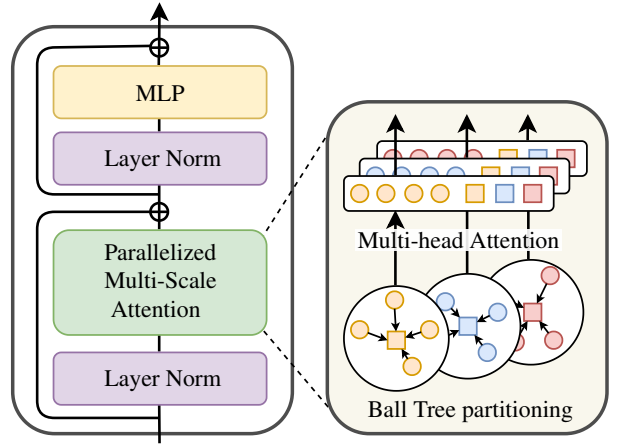

\paragraph{\METHOD\ Block}
Each \METHOD\ block is a modified pre-normalized Transformer block that operates on patch tokens and supernodes. Given point features $\mathbf{H} \in \mathbb{R}^{N \times C}$ and global supernodes $S \in \mathbb{R}^{KQ \times C}$:

\begin{enumerate}[leftmargin=20pt, topsep=0pt, itemsep=4pt]
\item \textbf{LayerNorm + PMSA.} We apply layer normalization to $\mathbf{H}$ (and $S$) and run PMSA across patches, with the global supernodes providing cross-patch context. This yields updated point features $\mathbf{H}'$ of shape $N \times C$ and updated supernodes $S'$. Both are added back to their inputs via residual connections.
\item \textbf{Feed-forward network.} A second LayerNorm followed by a shared MLP with nonlinearity (e.g., GELU) is applied pointwise to $\mathbf{H}'$, and its output is added residually, completing the standard Attention–FFN pattern.
\end{enumerate}

\vspace{0.2em}
We can write the $\ell$-th \METHOD\ block in compact form as
\begin{align*}
\widehat{\mathbf{H}}^{(\ell)},\, S^{(\ell)} &= \mathrm{PMSA}\big(\mathrm{LN}(\mathbf{H}^{(\ell-1)}),\, S^{(\ell-1)}\big) + \big(\mathbf{H}^{(\ell-1)},\, 0\big), \\
\mathbf{H}^{(\ell)} &= \mathrm{FFN}\big(\mathrm{LN}(\widehat{\mathbf{H}}^{(\ell)})\big) + \widehat{\mathbf{H}}^{(\ell)},
\end{align*}
with $\mathbf{H}^{(0)} = \mathbf{H}$ and $S^{(0)} = 0$. In the final block, the FFN is replaced by a task-specific head that maps hidden features to the target fields (e.g., pressure or velocity) at each point, while keeping the residual path. Stacking $B$ such blocks iteratively refines both point features and supernodes, with the choice of patches $K$ and supernodes $Q$ controlling the balance between receptive-field size and computational cost.

\section{Experiments}
\label{sec:experiments}

We evaluate the proposed \METHOD\ model on a wide range of benchmarks, including standard PDE problems and CFD tasks. We compare against state-of-the-art operator learning methods and attention-based models. Table~\ref{tab:datasets} provides a summary of the datasets, which span different geometry types, spatial dimensions, and problem sizes. Implementation details such as benchmarks details and hyper-parameters are given in Appendix~\ref{app:configurations}.

To evaluate the performance of the PMSA mechanism, we differentiate between \emph{single-branch} architectures and AB-UPT \citep{alkin2025abuptscalingneuralcfd}, which employs separate branches for geometry, surface, and volume inputs. For single-domain benchmarks that are not compatible with such a separation, we do not report AB-UPT results. 

\begin{table}[h]
    \caption{Summary of benchmark datasets used in our experiments. Each dataset is characterized by the geometry type, the physical problem dimension, and the average number of points.}
    \label{tab:datasets}
    \centering
    \resizebox{\linewidth}{!}{
            \begin{tabular}{llll}
                \toprule
                \textbf{Geometry} & \textbf{Benchmark} & \textbf{Dim} & \textbf{\#Nodes} \\
                \midrule
                Point Cloud & Elasticity~\cite{FNO,Li2022FourierNO} & 2D & 972 \\
                \midrule
                \multirow{3}{*}{Structured} & Plasticity~\cite{FNO,Li2022FourierNO} & 2D+Time & 3,131 \\
                & Airfoil~\cite{FNO,Li2022FourierNO} & 2D & 11,271 \\
                & Pipe~\cite{FNO,Li2022FourierNO} & 2D & 16,641 \\
                \midrule
                \multirow{2}{*}{Regular Grid} & Navier-Stokes~\cite{FNO,Li2022FourierNO} & 2D+Time & 4,096 \\
                & Darcy~\cite{FNO,Li2022FourierNO} & 2D & 7,225 \\
                \midrule
                \multirow{2}{*}{Unstructured} & ShapeNet Car~\cite{chang2015shapenet,umetani2018learning} & 3D & 32,186 \\
                & AhmedML~\cite{ashton2024ahmed} & 3D & $2\times10^{7}$ \\
                \bottomrule
            \end{tabular}
    }
\end{table}

\subsection{Standard PDE Benchmarks}
\label{sec:standard_pde_bench}

\textbf{Task} We consider six standard PDE surrogate benchmarks that are widely used in the operator learning literature: Elasticity, Plasticity, Airfoil, Pipe, Navier–Stokes, and Darcy. These benchmarks were introduced by FNO and geo-FNO~\cite{FNO,Li2022FourierNO} and have since become a standard suite for operator-learning work. They represent a broad spectrum of physical systems and computational challenges. Collectively, these benchmarks cover key PDE types: material stress/deformation (Elasticity, Plasticity), fluid dynamics governed by the Navier–Stokes equations (Airfoil, Pipe, and a spatio-temporal Navier–Stokes case), and porous-media flow described by Darcy’s law. For a brief description, Elasticity requires predicting internal stress fields in a solid object given its shape and boundary conditions. Plasticity involves forecasting the deformation of a metal under impact. Airfoil predicts the steady-state Mach number distribution around an airfoil for a given airflow condition. Pipe predicts fluid velocity through various 2D cross-sections of pipes. Navier–Stokes predicts the evolution of an incompressible fluid flow field for the next time steps, given the previous time steps. Darcy predicts pressure fields in subsurface flow through porous media. Details are given in Appendix~\ref{app:configurations}.

\textbf{Baselines} We compare \METHOD{} against baselines reported in \citet{wu2024Transolver} and \citet{zhdanov2025erwin}, spanning classical neural operators and transformer-based neural surrogates for PDEs. For all models, the relative $L_2$ error between predicted and ground-truth fields is reported. Results are averaged over all test samples. All experiments are conducted within the \textit{Neural-Solver-Library}\footnote{\href{https://github.com/thuml/Neural-Solver-Library}{https://github.com/thuml/Neural-Solver-Library}} framework, following the benchmarking setup of \citet{wu2024Transolver}. Transolver++ is excluded due to reproducibility challenges, which we discuss in detail in Appendix~\ref{app:app_reproducibility}.

\begin{table}[t]
    \caption{Performance comparison on standard benchmarks. Relative $L_2$ is recorded. A smaller value indicates better performance. For clarity, the best result is in bold and the second best is underlined. Promotion refers to the relative error reduction w.r.t.~the second best model ($1-\frac{\text{Our error}}{\text{The second best error}}$). “/” means that the baseline cannot apply to this benchmark. Baselines are taken from \citet{wu2024Transolver} and \citet{zhdanov2025erwin}.}
    \label{tab:mainres_standard}
    \centering
    \setlength{\tabcolsep}{3pt}
    \resizebox{\linewidth}{!}{
        \begin{tabular}{@{}l@{}cccccc@{}}
            \toprule
            \multirow{4}{*}{Model} & Point Cloud & \multicolumn{3}{c}{Structured Mesh} & \multicolumn{2}{c}{Regular Grid} \\
            \cmidrule(lr){2-2}\cmidrule(lr){3-5}\cmidrule(lr){6-7}
            & \multirow{2}{*}{Elasticity} & \multirow{2}{*}{Plasticity} & \multirow{2}{*}{Airfoil} & \multirow{2}{*}{Pipe} & Navier & \multirow{2}{*}{Darcy} \\
            & & & & & Stokes & \\
            \midrule
            FNO \citep{FNO} & / & / & / & / & 15.56 & 1.08 \\
            WMT \citep{Gupta2021MultiwaveletbasedOL} & 3.59 & 0.76 & 0.75 & 0.77 & 15.41 & 0.82 \\
            U-FNO \citep{Wen2021UFNOA} & 2.39 & 0.39 & 2.69 & 0.56 & 22.31 & 1.83 \\
            geo-FNO \citep{Li2022FourierNO} & 2.29 & 0.74 & 1.38 & 0.67 & 15.56 & 1.08 \\
            U-NO \citep{rahman2022u} & 2.58 & 0.34 & 0.78 & 1.00 & 17.13 & 1.13 \\
            F-FNO \citep{anonymous2023factorized} & 2.63 & 0.47 & 0.78 & 0.70 & 23.22 & 0.77 \\
            LSM \citep{wu2023LSM} & 2.18 & \underline{0.25} & 0.59 & 0.50 & 15.35 & \underline{0.65} \\
            \midrule
            Galerkin \citep{Cao2021ChooseAT} & 2.40 & 1.20 & 1.18 & 0.98 & 14.01 & 0.84 \\
            HT-Net \citep{liu2022htnet} & / & 3.33 & 0.65 & 0.59 & 18.47 & 0.79 \\
            OFormer \citep{li2023transformer} & 1.83 & 0.17 & 1.83 & 1.68 & 17.05 & 1.24 \\
            GNOT \citep{hao2023gnot} & 0.86 & 3.36 & 0.76 & 0.47 & 13.80 & 1.05 \\
            FactFormer \citep{li2023scalable} & / & 3.12 & 0.71 & 0.60 & 12.14 & 1.09 \\
            ONO \citep{anonymous2023improved} & 1.18 & 0.48 & 0.61 & 0.52 & 11.95 & 0.76 \\
            Transolver~\citep{wu2024Transolver} & 0.64 & \underline{0.12} & \underline{0.53} & \underline{0.33} & \underline{9.00} & \textbf{0.57} \\
            Erwin~\citep{zhdanov2025erwin} & \textbf{0.34} & \textbf{0.10} & 2.57 & 0.61 & N/A & N/A \\
            \midrule
            \textbf{\METHOD\ (Ours)} & \underline{0.48} & \textbf{0.10} & \textbf{0.51} & \textbf{0.31} & \textbf{6.32} & 0.63 \\
            \textit{Relative Promotion} & -41\% & 17\% & 4\% & 6\% & 30\% & -10\% \\
            \bottomrule
        \end{tabular}
    }\\[0pt]
    \raggedright\footnotesize{Values shown in units of $\times 10^{-2}$}
    \vspace{-5mm}
\end{table}

\textbf{Results} MSPT achieves state-of-the-art performance on four of six standard PDE benchmarks (Table~\ref{tab:mainres_standard}), consistently outperforming Transolver with substantial improvements on Navier–Stokes (30\%) and Elasticity (25\%). A key difference between the two models is how they combine information: Transolver compresses the domain into a fixed set of global slices and attends in that latent space, whereas \METHOD{} operates on local patches and summarizes them into supernodes. This preserves local detail and lets cross-patch interactions scale with the number of patches $K$, rather than through a fixed global bottleneck.
Compared to Erwin, \METHOD{} is weaker on Elasticity but clearly stronger on Airfoil. Erwin relies on strictly local attention, which is very efficient and works well when most interactions are local, but requires several layers to propagate information across the domain. In contrast, \METHOD{}’s supernodes provide a direct way to share information more broadly within each layer.

\begin{figure}[t]
    \centering

\newcommand{\imgheight}{2.1cm}
\newcommand{\titleheight}{1\baselineskip}

\begin{minipage}[t]{\linewidth}
    \centering
    \parbox[t][\titleheight][t]{\linewidth}{\centering \small \textbf{Plasticity}}\\[3pt]
    \begin{minipage}[t]{0.31\linewidth}
        \centering
        \includegraphics[height=\imgheight]{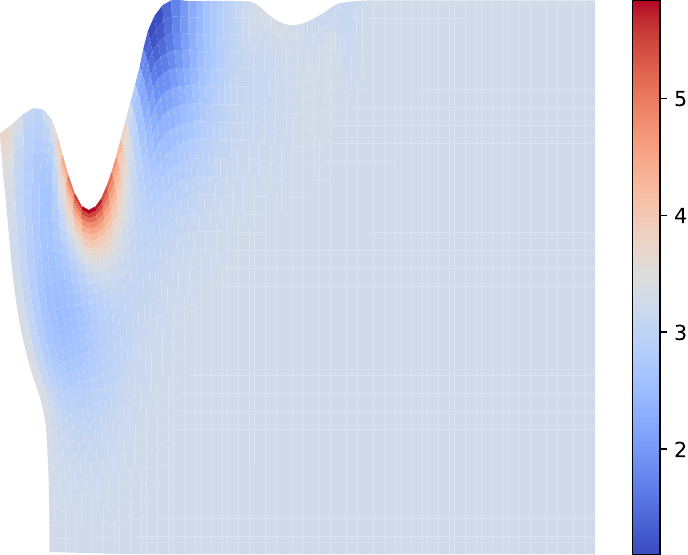}\\
    \end{minipage}\hfill
    \begin{minipage}[t]{0.31\linewidth}
        \centering
        \includegraphics[height=\imgheight]{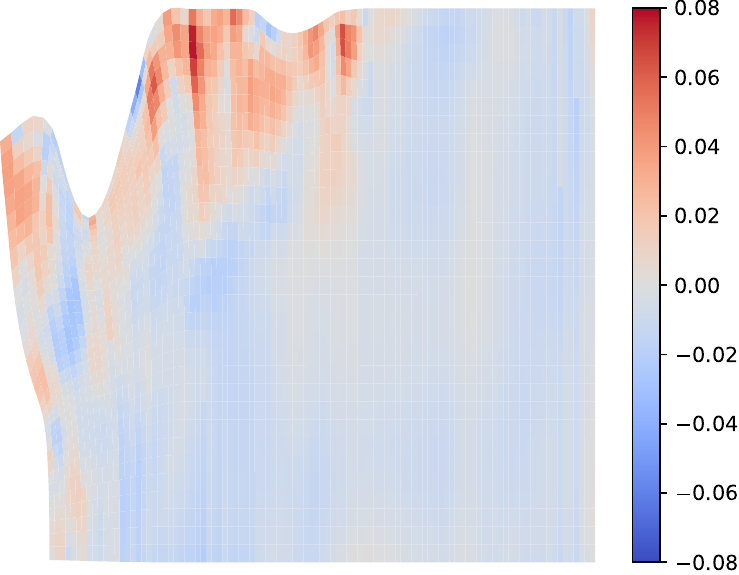}\\[2pt]
    \end{minipage}\hfill
    \begin{minipage}[t]{0.31\linewidth}
        \centering
        \includegraphics[height=\imgheight]{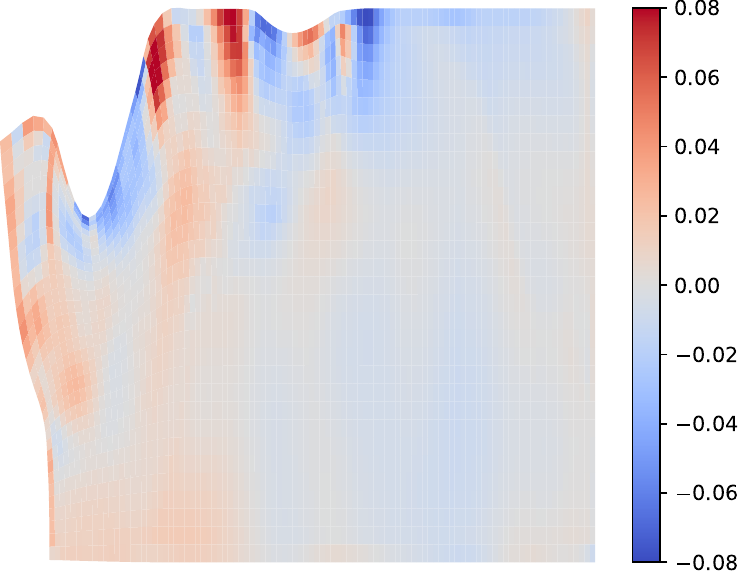}\\[2pt]
    \end{minipage}
\end{minipage}

\vspace{1em}

\begin{minipage}[t]{\linewidth}
    \centering
    \parbox[t][\titleheight][t]{\linewidth}{\centering \small \textbf{Pipe}}\\[3pt]
    \begin{minipage}[t]{0.31\linewidth}
        \centering
        \includegraphics[height=\imgheight]{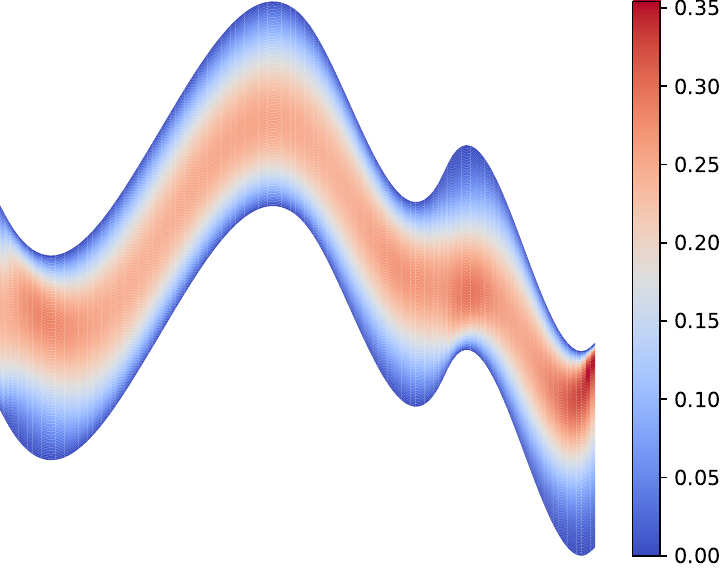}\\[2pt]
    \end{minipage}\hfill
    \begin{minipage}[t]{0.31\linewidth}
        \centering
        \includegraphics[height=\imgheight]{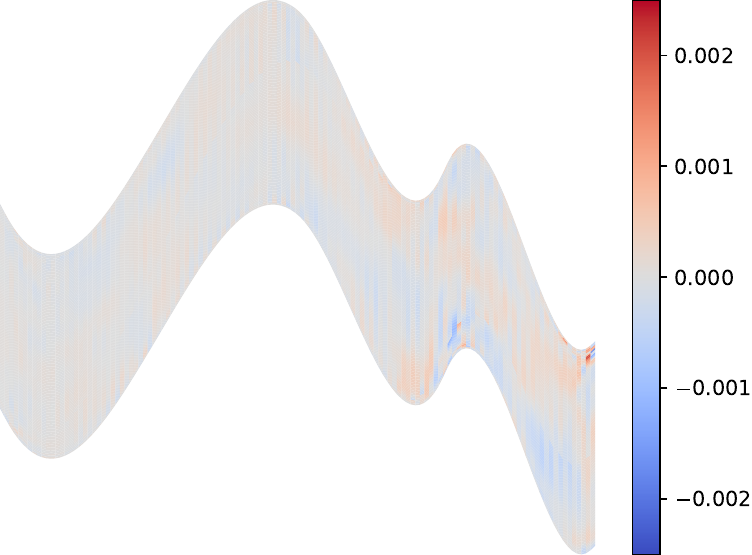}\\[2pt]
    \end{minipage}\hfill
    \begin{minipage}[t]{0.31\linewidth}
        \centering
        \includegraphics[height=\imgheight]{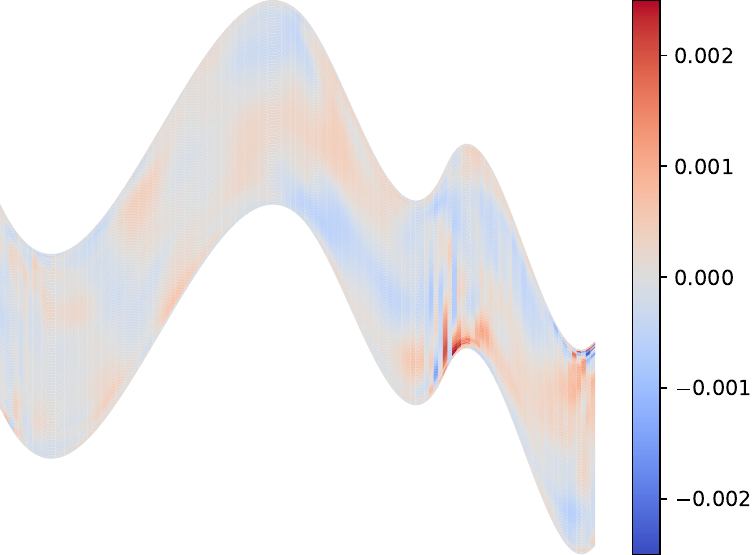}\\[2pt]
    \end{minipage}
\end{minipage}

\vspace{1em}

\begin{minipage}[t]{\linewidth}
    \centering
    \parbox[t][\titleheight][t]{\linewidth}{\centering \small \textbf{Navier-Stokes}}\\[3pt]
    \begin{minipage}[t]{0.31\linewidth}
        \centering
        \includegraphics[height=\imgheight]{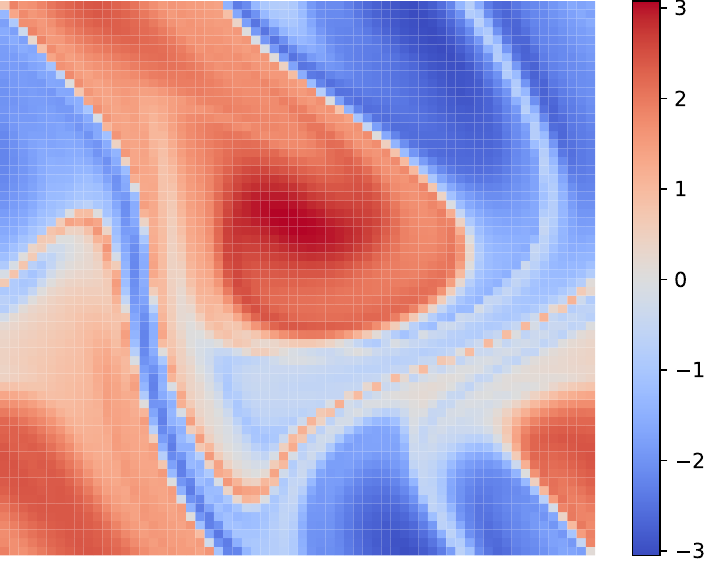}\\[2pt]
    \end{minipage}\hfill
    \begin{minipage}[t]{0.31\linewidth}
        \centering
        \includegraphics[height=\imgheight]{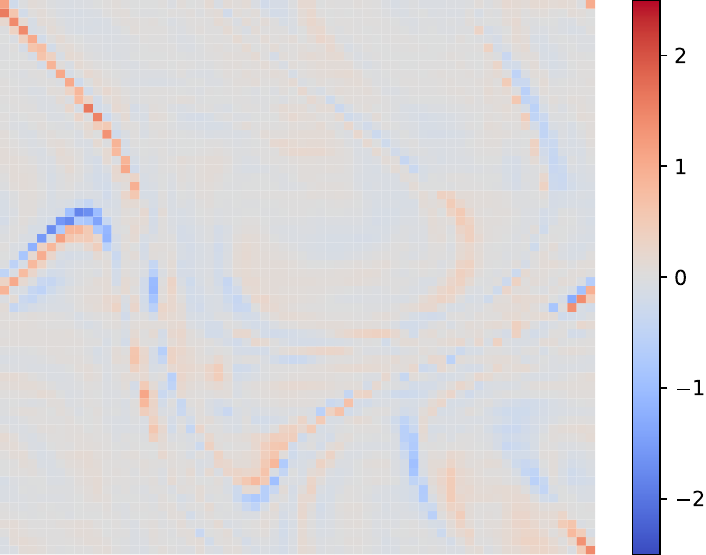}\\[2pt]
    \end{minipage}\hfill
    \begin{minipage}[t]{0.31\linewidth}
        \centering
        \includegraphics[height=\imgheight]{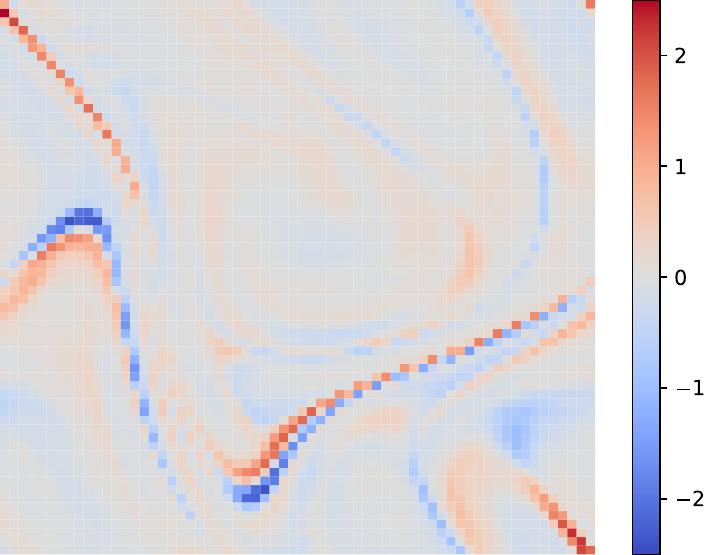}\\[2pt]
    \end{minipage}
\end{minipage}

\vspace{1em}

\begin{minipage}[t]{\linewidth}
    \centering
    \parbox[t][\titleheight][t]{\linewidth}{\centering \small \textbf{ShapeNet Car}}\\[3pt]
    \begin{minipage}[t]{0.31\linewidth}
        \centering
        \includegraphics[height=\imgheight]{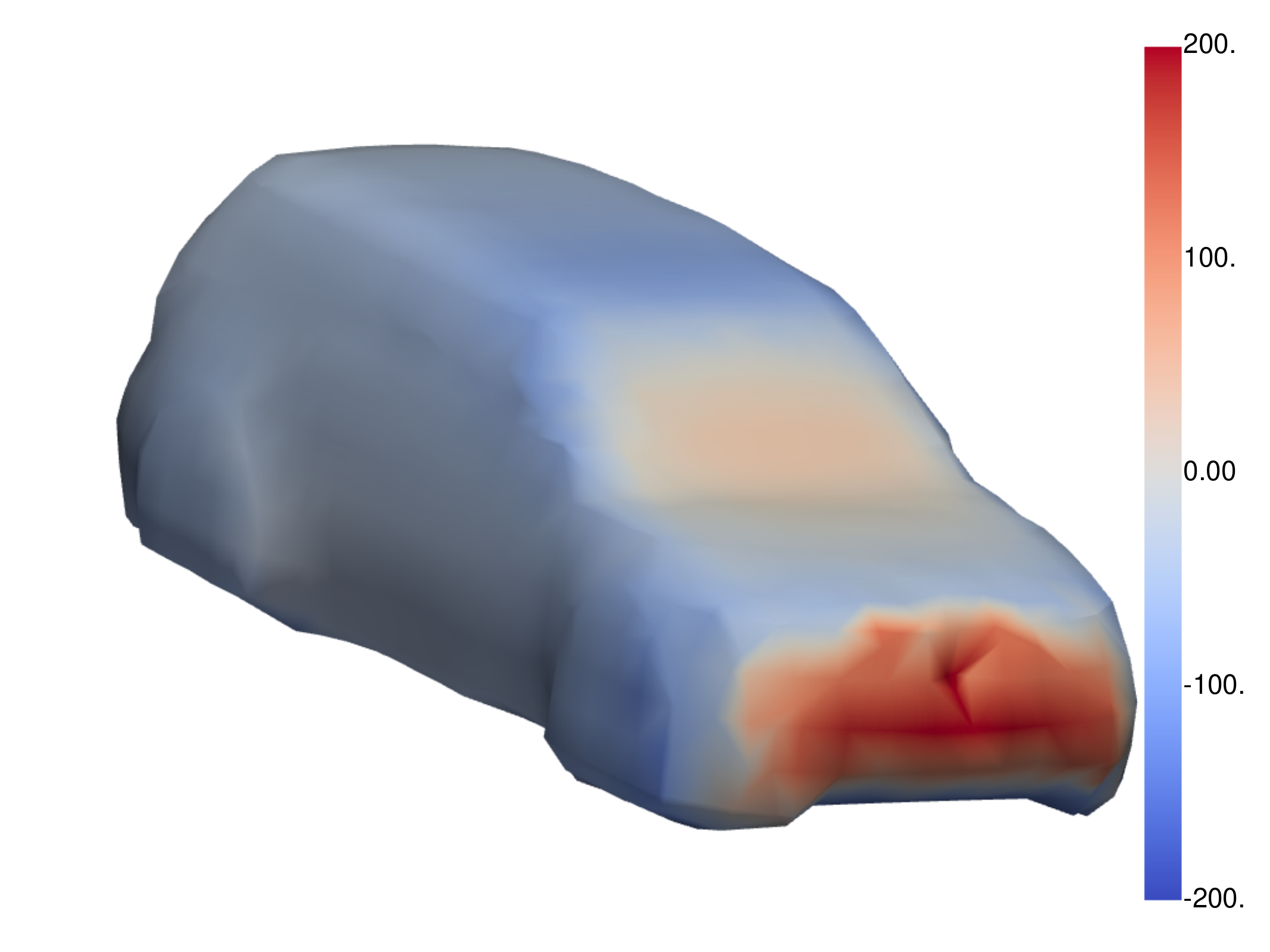}\\[2pt]
        \hspace{-0.5em}\small Ground-truth
    \end{minipage}\hfill
    \begin{minipage}[t]{0.31\linewidth}
        \centering
        \includegraphics[height=\imgheight]{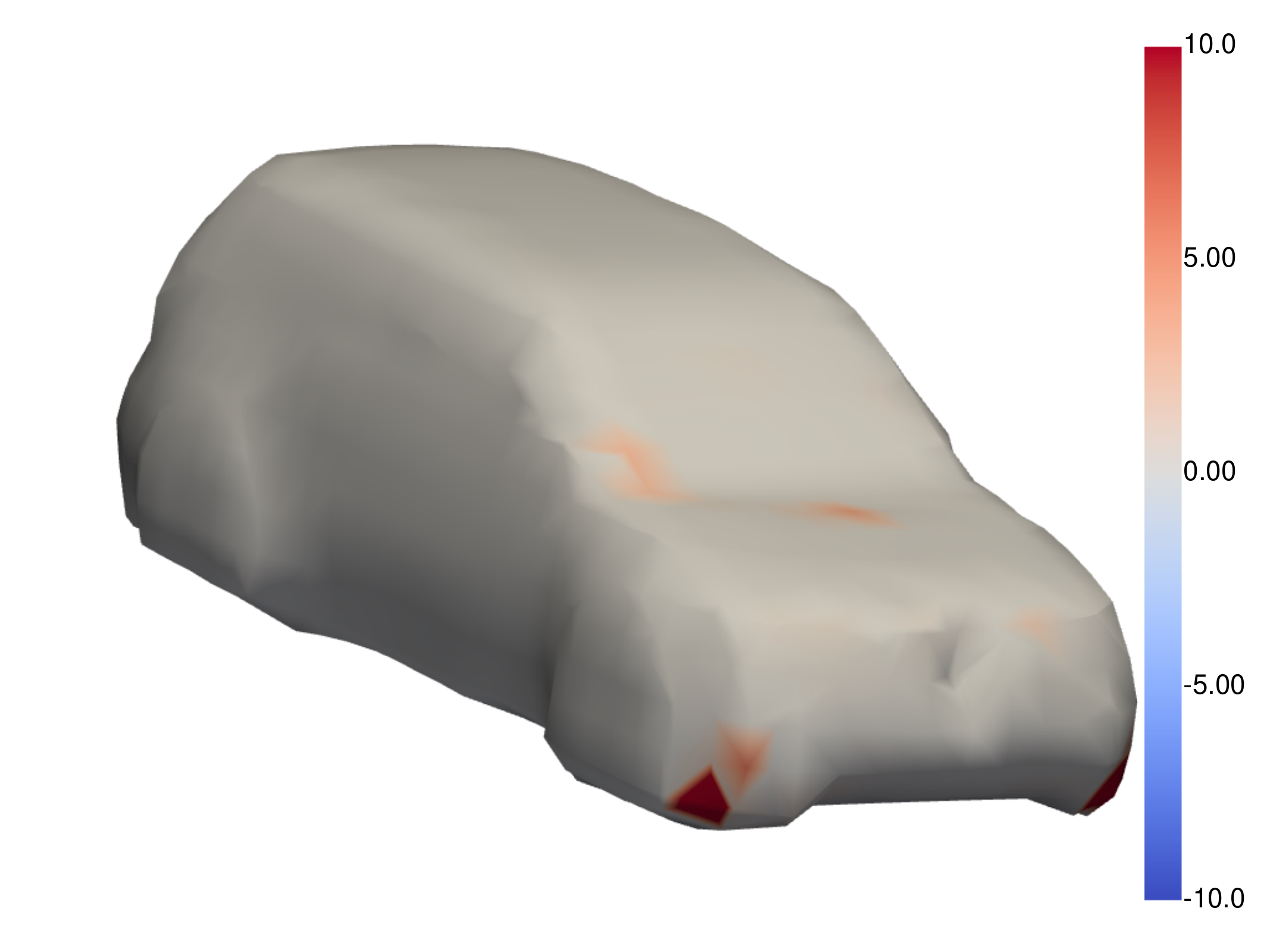}\\[2pt]
        \hspace{-0.5em}\small \METHOD{} (Ours)
    \end{minipage}\hfill
    \begin{minipage}[t]{0.31\linewidth}
        \centering
        \includegraphics[height=\imgheight]{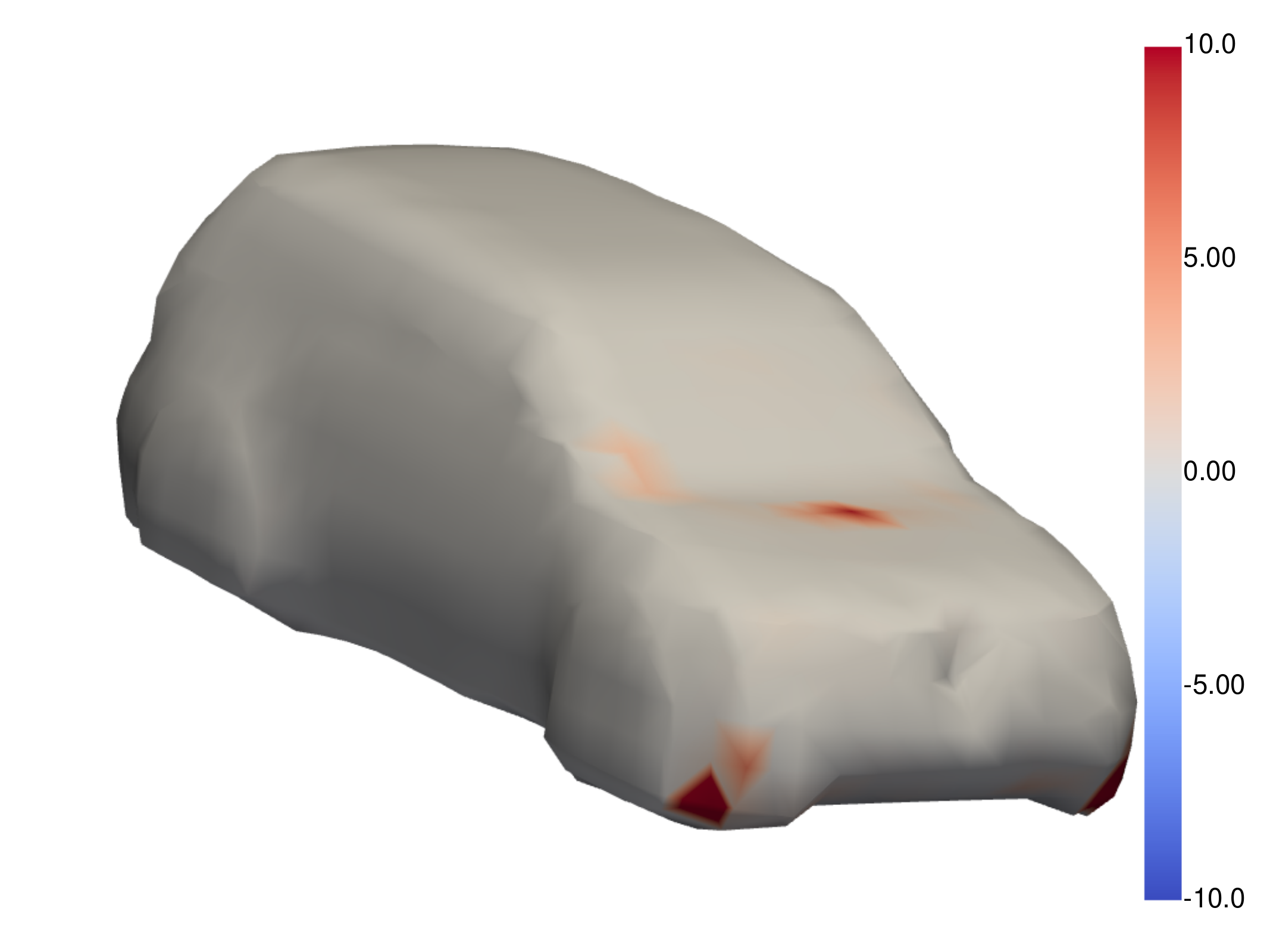}\\[2pt]
        \hspace{-0.5em}\small Transolver
    \end{minipage}
\end{minipage}

    \caption{Examples of relative L2 error maps for the Pipe, Navier-Stokes and ShapeNet Car datasets. For ShapeNet Car we show surface-pressure errors. 
    See Appendix~\ref{app:app_pde_benchmark} for more visualizations.}
    \label{fig:pde_benchmarks}
\end{figure}

\subsection{CFD Benchmarks}

We also evaluate our model on two design-oriented CFD benchmarks, which involve learning aerodynamic flow fields around complex 3D shapes: ShapeNet-Car and AhmedML. These problems are significantly larger in scale and reflect practical design optimization settings.

\subsubsection{ShapeNet-Car}
\label{sec:shapenet-car}

\textbf{Task} 
The ShapeNet-Car benchmark focuses on 3D aerodynamic field reconstruction and drag coefficient prediction for automotive shapes. Given a triangulated car surface mesh from the ShapeNet dataset, the model is required to predict steady-state velocity and pressure fields in the surrounding fluid volume, as well as the pressure distribution on the car’s surface. These predicted fields enable the computation of coefficients such as the aerodynamic drag coefficient $C_D$ using standard formulas.

\textbf{Baselines}
We compare \METHOD{} against graph-based networks, neural operators for unstructured 3D domains, transformer-based CFD surrogates, and AB-UPT~\citep{alkin2025abuptscalingneuralcfd}, under the evaluation setup of \citet{wu2024Transolver}. Metrics include relative $L_2$ on volume and surface fields, drag error $C_D$, and Spearman’s $\rho_D$ for ranking quality. All experiments are run within the \textit{Neural-Solver-Library}\footnote{\href{https://github.com/thuml/Neural-Solver-Library}{https://github.com/thuml/Neural-Solver-Library}} framework, following the training setup of \citet{wu2024Transolver}. For AB-UPT, we use the official code release\footnote{\href{https://github.com/Emmi-AI/anchored-branched-universal-physics-transformers.git}{https://github.com/Emmi-AI/anchored-branched-universal-physics-transformers.git}} but note that the original training pipeline is not publicly available; we therefore reproduce its results using the configurations reported by \citet{alkin2025abuptscalingneuralcfd} within the Neural-Solver-Library setup. We report both the original AB-UPT numbers from their paper and our reproduction. Discrepancies are analyzed in Appendix~\ref{app:app_reproducibility}.

\textbf{Results}
On ShapeNet-Car (Table~\ref{tab:design_tasks}), \METHOD{} is the best single-branch model, with improvements over Transolver on both volume fields and drag. This matches the behavior seen on the standard PDE benchmarks: Transolver aggregates information into a fixed set of global slices, whereas \METHOD{} operates on local patches and uses supernodes for global communication, preserving local detail while avoiding a hard global bottleneck and oversmoothing. AB-UPT, which introduces separate surface and volume branches specialized for CFD field reconstruction, attains lower field errors in the original report but yields weaker design metrics in our reproduction (Appendix~\ref{app:app_reproducibility}). The same branching idea can be applied to \METHOD{}, yielding a dual-branch variant while preserving its coupled local–global communication (Section~\ref{sec:mspt}).

\begin{table}
    \caption{Performance on ShapeNet Car. Relative $L_2$ for volume / surface fields and drag coefficient; Spearman $\rho_D$ for ranking quality. Baselines are taken from \citet{wu2024Transolver}, except for AB-UPT, which is taken from \citet{alkin2025abuptscalingneuralcfd}. \METHOD{} is compared primarily against one-branched models (e.g., Transolver, GNOT, GINO), while AB-UPT is shown for reference as a two-branched specialized CFD architecture. Promotion refers to the relative error reduction w.r.t.~the second best model ($1-\frac{\text{Our error}}{\text{Second best error}}$).}
    \label{tab:design_tasks}
    \centering
    \setlength{\tabcolsep}{4pt}
    \resizebox{\linewidth}{!}{
    \begin{tabular}{@{}lcccc@{}}
        \toprule
        \multirow{2}{*}{\textbf{Model}} & \multicolumn{4}{c}{\textbf{ShapeNet Car}} \\
        \cmidrule(lr){2-5}
        & Volume $\downarrow$ & Surf $\downarrow$ & $(C_D)$ $\downarrow$ & $(\rho_D)$ $\uparrow$ \\
        \midrule
        Simple MLP & 5.12 & 13.04 & 3.07 & 94.96 \\
        GraphSAGE~\citep{hamilton2017inductive} & 4.61 & 10.50 & 2.70 & 96.95 \\
        PointNet~\citep{qi2017pointnet} & 4.94 & 11.04 & 2.98 & 95.83 \\
        Graph U-Net~\citep{gao2019graph} & 4.71 & 11.02 & 2.26 & 97.25 \\
        MeshGraphNet~\citep{pfaff2021learning} & 3.54 & 7.81 & 1.68 & 98.40 \\
        \midrule
        GNO~\citep{Kovachki2023NeuralOL} & 3.83 & 8.15 & 1.72 & 98.34 \\
        GALERKIN~\citep{Cao2021ChooseAT} & 3.39 & 8.78 & 1.79 & 97.64 \\
        GEO-FNO~\citep{Li2022FourierNO} & 16.70 & 23.78 & 6.64 & 82.80 \\
        GNOT~\citep{hao2023gnot} & 3.29 & 7.98 & 1.78 & 98.33 \\
        GINO~\citep{li2023geometryinformed} & 3.86 & 8.10 & 1.84 & 98.26 \\
        3D-GEOCA~\citep{anonymous2023geometryguided} & 3.19 & 7.79 & 1.59 & 98.42 \\
        Transolver~\citep{wu2024Transolver} & \underline{2.07} & \underline{7.45} & \underline{1.03} & \underline{99.35} \\
        \midrule
        \textbf{\METHOD\ (Ours)} & \textbf{1.89} & \textbf{7.41} & \textbf{0.98} & \textbf{99.41} \\
        \textit{Relative Promotion} & +8.7\% & +0.5\% & +4.9\% & +0.06\% \\
        \midrule
        AB-UPT~\citep{alkin2025abuptscalingneuralcfd} & \textbf{1.16} & \textbf{4.81} & N/A & N/A \\
        AB-UPT~\citep{alkin2025abuptscalingneuralcfd} (\textit{repr.}) & 2.51 & 7.67 & 2.20 & 97.48 \\
        \bottomrule
    \end{tabular}
    }\\[0pt]
    \raggedright\footnotesize{Values shown in units of $\times 10^{-2}$}\\
    \footnotesize{Differences in AB-UPT~\citep{alkin2025abuptscalingneuralcfd} results are discussed in Appendix~\ref{app:app_reproducibility}.}
    \vspace{-4mm}
\end{table}

\subsubsection{AhmedML}

\textbf{Task}
This benchmark targets 3D aerodynamic field reconstruction and force-coefficient estimation for the Ahmed body family. Given the body surface mesh, the model must predict velocity and pressure throughout the surrounding domain and the surface pressure on the body.

\textbf{Baselines}
We follow the evaluation setup of \citet{alkin2025abuptscalingneuralcfd}, comparing \METHOD{} against PointNet~\citep{qi2017pointnet}, Graph U-Net~\citep{gao2019graph}, GINO~\citep{li2023geometryinformed}, LNO~\citep{wang2024LNO}, OFormer~\citep{li2023transformer}, Transolver~\citep{wu2024Transolver}, UPT~\citep{alkin2024upt}, and AB-UPT~\citep{alkin2025abuptscalingneuralcfd}. We adopt the experimental setup of \citet{alkin2025abuptscalingneuralcfd} and use the official AB-UPT code implementation \footnote{\href{https://github.com/Emmi-AI/anchored-branched-universal-physics-transformers.git}{https://github.com/Emmi-AI/anchored-branched-universal-physics-transformers.git}} to train our model. However, we note that the original training pipeline used in AB-UPT is not open-sourced. Therefore, to ensure a fair comparison, we reproduce the experiment using the shared training routine and report our results alongside those reported by the authors.

\textbf{Results}
On AhmedML (Table~\ref{tab:design_tasks_ahmed}), \METHOD{} is the best single-branch model, with clear improvements over Transolver on both volume and surface fields. This aligns with the trends on the standard PDE benchmarks and ShapeNet-Car (Sections~\ref{sec:standard_pde_bench} and~\ref{sec:shapenet-car}), where PMSA gives a more effective local–global interaction than Transolver’s fixed set of global slices. AB-UPT, which adds separate surface and volume branches specialized for CFD field reconstruction, achieves lower field errors in the original report but yields worse metrics in our reproduction (Appendix~\ref{app:app_reproducibility}). The same branching idea can, in principle, be applied to \METHOD{}, yielding a dual-branch variant while preserving its coupled local–global communication (Section~\ref{sec:mspt}).

\begin{table}[t]
    \caption{Performance on AhmedML. Relative $L_2$ for volume and surface fields. Baselines are taken from \citet{alkin2025abuptscalingneuralcfd}. \METHOD{} is compared primarily against one-branched models, while AB-UPT is shown for reference as a two-branched specialized CFD architecture. Promotion refers to the relative error reduction w.r.t.~the second best model ($1-\frac{\text{Our error}}{\text{The second best error}}$).}
    \label{tab:design_tasks_ahmed}
    \centering
    \setlength{\tabcolsep}{4pt}
    \resizebox{0.65\linewidth}{!}{
    \begin{tabular}{@{}lcc@{}}
        \toprule
        \multirow{2}{*}{\textbf{Model}} & \multicolumn{2}{c}{\textbf{AhmedML}} \\
        \cmidrule(lr){2-3}
        & Volume $\downarrow$ & Surf $\downarrow$ \\
        \midrule
        PointNet~\citep{qi2017pointnet} & 5.44 & 8.02 \\
        Graph U-Net~\citep{gao2019graph} & 4.15 & 6.46 \\
        GINO~\citep{li2023geometryinformed} & 6.23 & 7.90 \\
        LNO~\citep{wang2024LNO} & 7.59 & 12.95 \\
        UPT~\citep{alkin2024upt} & 2.73 & 4.25 \\
        OFormer~\citep{li2023transformer} & 3.63 & 4.12 \\
        Transolver~\citep{wu2024Transolver} & \underline{2.05} & \underline{3.45} \\
        Transformer~\citep{li2023transformer} & 2.09 & 3.41 \\
        \midrule
        \textbf{\METHOD\ (Ours)} & \textbf{2.04} & \textbf{3.22} \\
        \textit{Relative Promotion} & +0.49\% & +6.67\% \\
        \midrule
        AB-UPT~\citep{alkin2025abuptscalingneuralcfd} & \textbf{1.90} & \textbf{3.01} \\
        AB-UPT~\citep{alkin2025abuptscalingneuralcfd} (\textit{repr.}) & 2.39 & 4.33 \\
        \bottomrule
    \end{tabular}
    }\\[0pt]
    \raggedright\footnotesize{Values shown in units of $\times 10^{-2}$}\\
    \footnotesize{Differences in AB-UPT~\citep{alkin2025abuptscalingneuralcfd} results are discussed in Appendix~\ref{app:app_reproducibility}.}
    \vspace{-4mm}
\end{table}

\subsection{Hyperparameter and Ablation Studies}

We conducted a series of hyperparameter and ablation studies on the ShapeNet-Car benchmark to evaluate the impact of key design choices in the \METHOD\ architecture.

\paragraph{Performance vs. Number of Patches ($K$)}
We analyze the effect of the patch count $K$ under ball-tree partitioning on ShapeNet-Car (Table~\ref{tab:shapenet_balltree_k}). The relationship between $K$ and test loss is non-monotonic. With few patches ($K=32$), each patch is large ($L = N/K \sim 1000$ points in ShapeNet-Car), providing substantial local context but limiting global communication through the supernodes. As $K$ increases, patches become smaller, initially increasing test loss due to reduced local context. Beyond a threshold ($K \geq 512$ in this study), performance improves again as the larger number of supernodes enhances global interactions and enables more fine-grained modeling. This behavior reflects a fundamental trade-off: too few patches oversmooth local details, while too many fragment long-range coherence. An intermediate $K$ balances these competing demands, optimizing both local resolution and global context (Appendix~\Cref{app:chooseK}).

\begin{figure}[ht]
    \centering
    \includegraphics[width=0.855\linewidth]{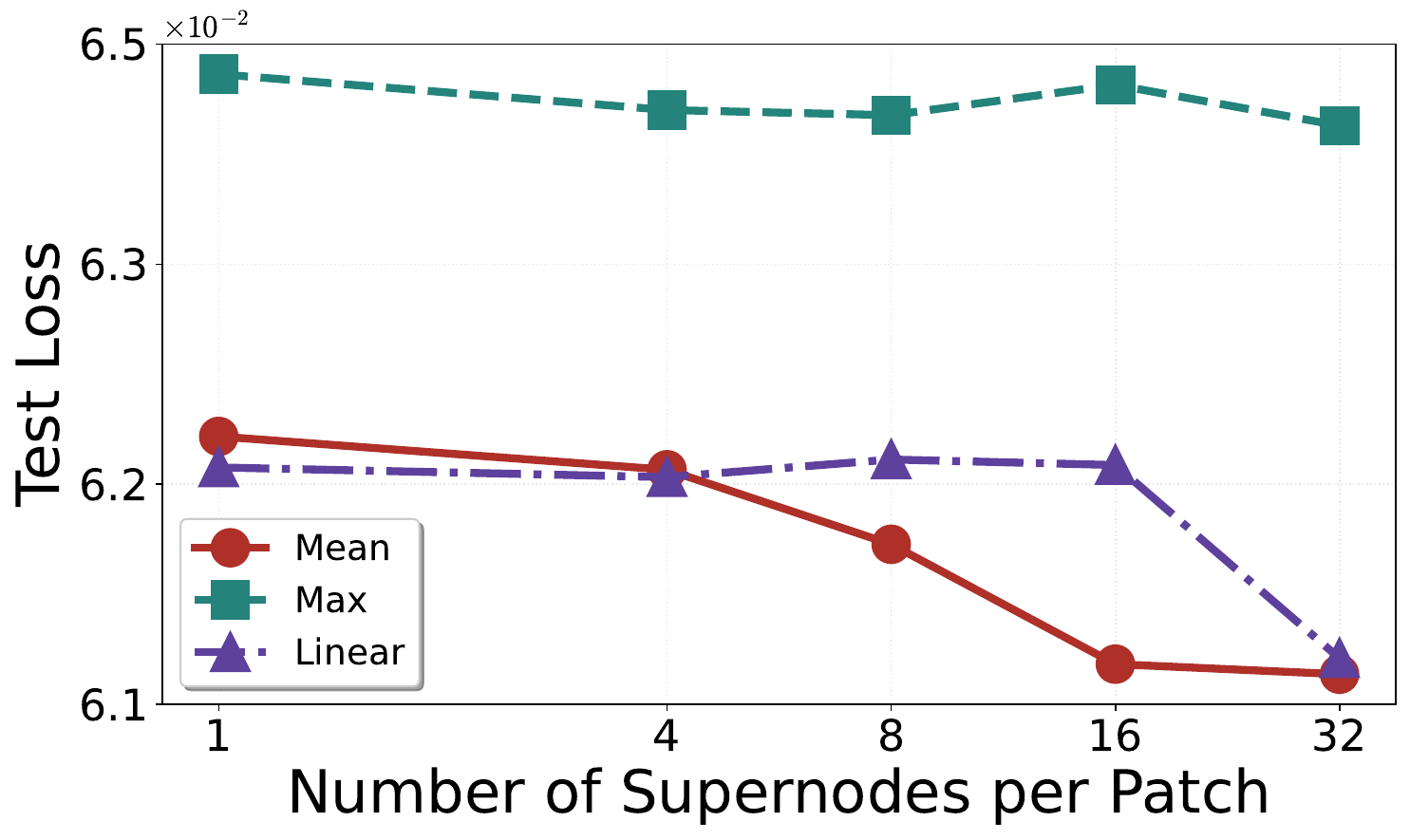}
    \caption{Study of pooling method and number of supernode tokens $Q$ per patch. We report the \emph{test loss}, defined as $\mathcal{L}=\mathcal{L}_v + 0.5\,\mathcal{L}_s$ (volume and surface terms), following the benchmark protocol (see \ref{app:impl_details}). Mean pooling consistently outperforms max pooling and learned linear projection for various $Q$ values. Increasing $Q$ generally lowers the validation loss.}
    \label{fig:ablation_pooling}
\end{figure}

\begin{table}[ht]
    \caption{ShapeNet-Car test loss (see Appendix~\ref{app:configurations}) as a function of the number of patches $K$.}
    \label{tab:shapenet_balltree_k}
    \centering
    \setlength{\tabcolsep}{4pt}
    \resizebox{0.8\linewidth}{!}{
    \begin{tabular}{@{}lcccccc@{}}
        \toprule
        \multirow{2}{*}{} & \multicolumn{6}{c}{\textbf{Number of patches $K$}} \\
        \cmidrule(lr){2-7}
        & 32 & 64 & 128 & 256 & 512 & 1024 \\
        \midrule
        Test Loss & 6.08 & 6.23 & 6.83 & 6.77 & 6.37 & 5.99 \\
        \bottomrule
    \end{tabular}
    }\\[0pt]
    \raggedright\footnotesize{Values shown in units of $\times 10^{-2}$.\par}
    \vspace{-4mm}
\end{table}
\paragraph{Pooling Method and Number of Supernodes}
We investigated the choice of pooling operator and the number of supernode tokens $Q$ per patch through three pooling types: mean, top-$Q$, and a learned linear projection. The learned projection was implemented as a fully-connected layer mapping $\mathbb{R}^{L \times F} \to \mathbb{R}^{Q \times F}$, with $Q$ ranging from 1 to 32. As shown in Figure~\ref{fig:ablation_pooling}, mean pooling consistently achieved good performance. In contrast, max pooling performed worse, emphasizing extreme values that may not represent typical patch characteristics. The learned projection, despite being flexible, did not consistently outperform mean pooling.

Regarding the number of supernodes, employing multiple tokens per patch ($Q > 1$) generally improved performance by enabling richer patch representation. Specifically, increasing $Q$ allows to capture fine-grained information within each patch. However, large $Q$ values reduce the computational benefits of pooling while adding additional overhead.

\subsection{Efficiency Analysis}
\label{sec:efficiency_analysis}
A primary motivation for \METHOD\ is scalability to large point sets. We analyze computational and memory scaling, measuring peak GPU memory usage and runtime for varying point ($N$) and patch ($K$) counts.

\begin{figure}[t]
    \centering
    \includegraphics[width=0.855\linewidth]{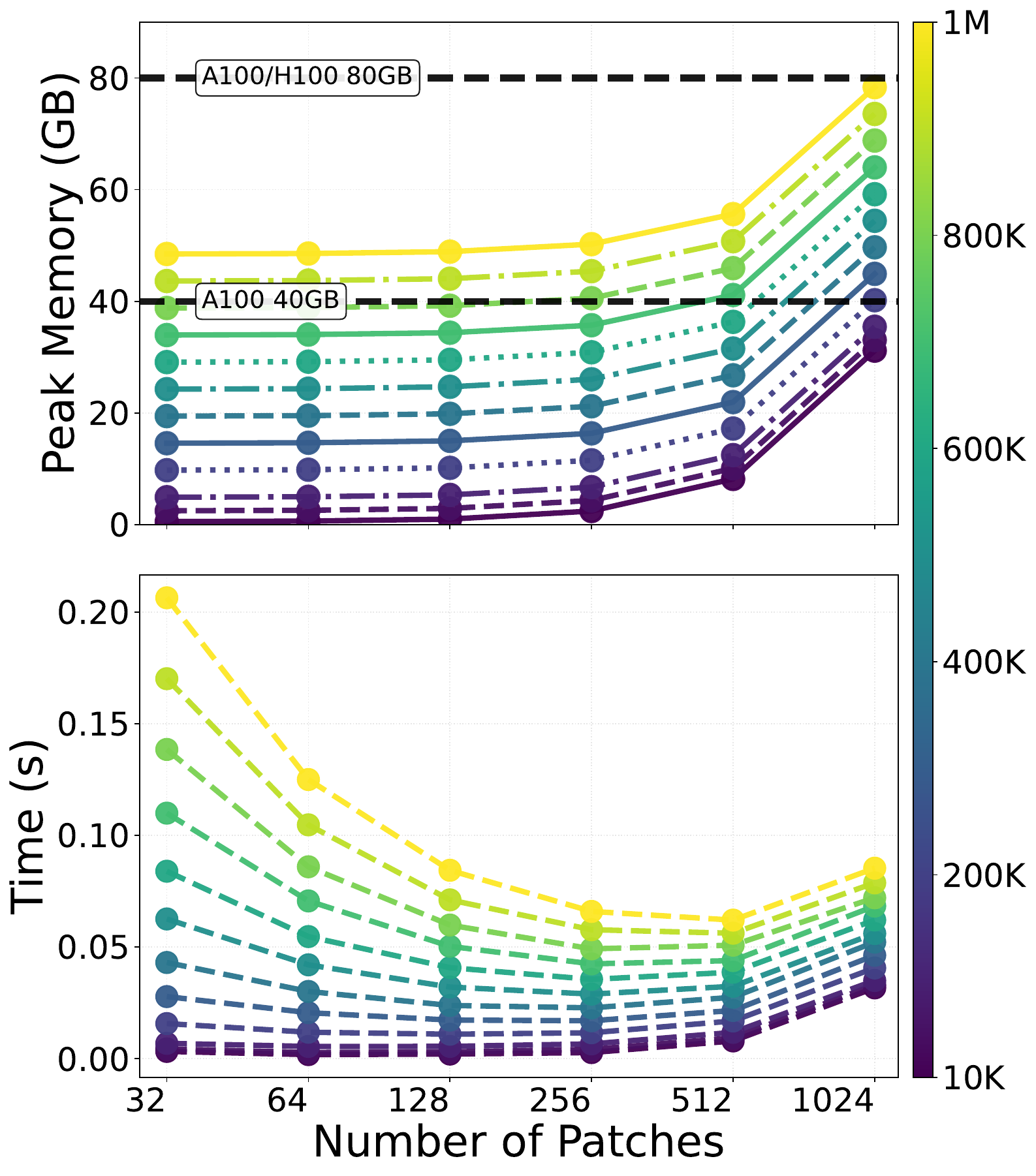}
    \caption{Peak GPU memory usage (\textit{top}) and wall-clock runtime per forward pass (\textit{bottom}) as a function of the number of patches, across several input resolutions. Runtime is measured end-to-end and includes preprocessing (ball-tree construction, permutation, and padding). Colors correspond to the input resolution (total number of points), as indicated by the color bar.}
    \label{fig:memory_vs_V}
    \vspace{-5mm}
\end{figure}

Figure~\ref{fig:memory_vs_V} shows peak memory usage increases almost linearly with the number of input points. For $V\!\in\![64,256]$, the model remains below the A100's 40 GB memory limit up to approximately $800$k points and stays under 80 GB at one million points. Figure~\ref{fig:memory_vs_V} illustrates the trade-off for $K$: very few patches (32) minimize overhead but result in slower execution, while fine partitions (512--1024) significantly increase memory usage (e.g., $>30$ GB at $10$k points). Across parameter sweeps, $K=128$ provides the optimal balance for the ShapeNet-Car benchmark, enabling a million-point batch in $0.084$ s under 50 GB. This demonstrates how parallelized multi-scale attention can be optimized for hardware constraints without sacrificing throughput.

\section{Conclusion and Future Work}
\label{sec:conclusion}

This paper introduces the Multi-Scale Patch Transformer (MSPT) for solving PDEs on arbitrary geometries. MSPT captures both fine-grained local interactions and long-range global dependencies via Parallelized Multi-Scale Attention (PMSA), which partitions the domain into spatial patches with ball trees and, in a single operation, performs local self-attention within patches while applying global attention to pooled representations. Experimental results show state-of-the-art accuracy on standard PDE benchmarks and CFD design tasks. Our analysis also demonstrates favorable memory and runtime scaling, with competitive throughput at million-point resolutions on a single GPU. Future work includes a branched MSPT variant for specialized surface/volume coupling, and improved pooling and partitioning strategies to further enhance global communication.

\newpage
\section*{Acknowledgments}

This project was supported by the ELLIS Unit Amsterdam and carried out using the Dutch national e-infrastructure, with the support of SURF through the use of the Snellius supercomputer. JWvdM acknowledges support from the European Union Horizon Framework Programme (Grant agreement ID: 101120237)

{
    \small
    \bibliographystyle{ieeenat_fullname}
    \bibliography{main}
}

\newpage
\appendix
\onecolumn
\section{Implementation}
\label{app:configurations}

\subsection{Balltree Partitioning}
\label{app:balltree}

\paragraph{Ball trees.}
A ball tree is a hierarchical spatial partitioning data structure that recursively decomposes a point set into nested metric balls. Given a set of points $\mathcal{P} \subset \mathbb{R}^D$, each node $u$ in the tree stores a subset $\mathcal{P}_u \subseteq \mathcal{P}$ together with a center $\mathbf{c}_u \in \mathbb{R}^D$ and radius $r_u > 0$ such that
\[
\mathcal{P}_u \subseteq B(\mathbf{c}_u, r_u)
:= \{\mathbf{x} \in \mathbb{R}^D : \|\mathbf{x} - \mathbf{c}_u\|_2 \le r_u\}.
\]
The root node contains all points. Internal nodes recursively split their point set into two (approximately) balanced children, each associated with a tighter enclosing ball, until a prescribed leaf capacity is reached.

\paragraph{Construction.}
In our implementation we build a balanced binary ball tree from the point coordinates using the public \texttt{balltree-erwin}~\citep{zhdanov2025erwin}. Starting from the full set $\mathcal{P}$, each internal node $u$ is constructed by:
\begin{enumerate}
    \item choosing a splitting direction based on a pair of distant points in $\mathcal{P}_u$;
    \item partitioning $\mathcal{P}_u$ into two subsets of roughly equal size along this direction;
    \item fitting enclosing balls for each child subset by setting the center to the mean and the radius to the maximal distance to that mean.
\end{enumerate}
This produces a tree of depth $O(\log N)$ with $O(N)$ nodes, and the overall construction cost is $O(N \log N)$.

\paragraph{Leaf ordering and patch extraction.}
To obtain patches for PMSA, we traverse the leaves of the ball tree in depth-first order and record the points in that leaf order. This induces a permutation $\pi$ of $\{1,\dots,N\}$ such that nearby indices in the permuted sequence correspond to spatially close points. We then form patches by taking contiguous blocks of length $L$ in this ordered sequence:
\[
\mathcal{P}_k = \big\{ \mathbf{p}_{\pi((k-1)L+1)}, \dots, \mathbf{p}_{\pi(kL)} \big\}, 
\qquad k=1,\dots,K,
\]
with padding applied if $N$ is not divisible by $L$. In this view, each patch can be interpreted as a collection of leaves and small subtrees that occupy a localized region of the domain. Once the permutation $\pi$ is fixed, all subsequent transformer blocks operate on the reordered sequence and can use simple linear splitting into contiguous patches.

\paragraph{Reuse across blocks.}
We emphasize that the ball tree is constructed \emph{once} per input sample, based solely on the initial coordinates. The induced permutation and patch layout are reused across all \METHOD{} blocks. This avoids the cost of rebuilding trees at each layer while preserving spatial locality throughout the network. For inputs that already come with a natural ordering on a regular grid, we skip the ball-tree construction and directly apply linear partitioning, which can be viewed as a degenerate case of ball-tree partitioning.

\begin{figure}[t]
    \centering
    \includegraphics[width=0.5\linewidth]{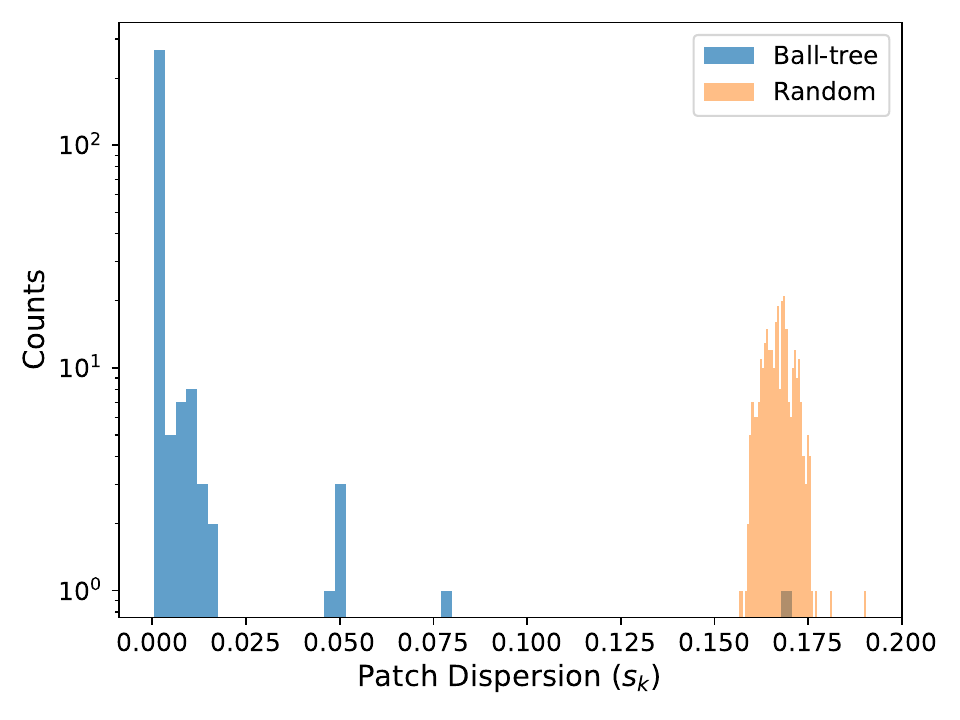}
    \caption{Histogram of per-patch spatial dispersion $s_k$ on Darcy after ball-tree DFS ordering and contiguous patch extraction ($K=300$; log-scaled y-axis). Ball-tree ordering yields predominantly compact patches (mass concentrated at low dispersion) with a small outlier tail, while a random permutation produces substantially larger dispersion (Table~\ref{tab:patch_dispersion_stats}).}
    \label{fig:patch_dispersion_hist}
\end{figure}

\paragraph{Patch locality diagnostic.}
To assess whether depth-first leaf traversal introduces occasional non-local ``jumps'' that mix distant points within a patch, we measure the spatial dispersion of each extracted patch after applying the ball-tree-induced permutation.
For a patch $\mathcal{P}_k$ with points $\{x_i\}_{i\in\mathcal{P}_k}$, we compute
\[
s_k=\frac{1}{|\mathcal{P}_k|}\sum_{i\in\mathcal{P}_k}\lVert x_i-\bar{x}_k\rVert_2^2,
\qquad
\bar{x}_k=\frac{1}{|\mathcal{P}_k|}\sum_{i\in\mathcal{P}_k} x_i,
\]
(i.e., the trace of the within-patch coordinate covariance / mean squared radius), and plot a histogram of $\{s_k\}_{k=1}^K$.
For example, in the Darcy benchmark (with $K=300$ patches), Figure~\ref{fig:patch_dispersion_hist} shows that the patch-dispersion distribution under ball-tree DFS ordering is concentrated at low values with only a small outlier tail, indicating that the induced permutation yields predominantly spatially coherent patches in practice; in contrast, a random permutation produces substantially larger dispersion (Table~\ref{tab:patch_dispersion_stats}).
Quantitatively, Table~\ref{tab:patch_dispersion_stats} reports the median / p90 / p99 / max of $s_k$ across benchmarks and shows substantially smaller dispersion under ball-tree ordering than under a random permutation (e.g., Elasticity: median $0.0065$ vs.\ $0.2035$; Darcy: median $0.0012$ vs.\ $0.1672$).

\begin{table}[ht]
\centering
\caption{Patch-dispersion statistics across PDE benchmarks (mean squared radius within each patch; lower is better). We report the median, 90th percentile (p90), 99th percentile (p99), and maximum over patches for the ball-tree DFS ordering versus a random permutation, using the same number of patches $K$ as in our main experiments.}
\label{tab:patch_dispersion_stats}
\setlength{\tabcolsep}{4.5pt}
\renewcommand{\arraystretch}{1.05}
\begin{tabular}{l c r r r r r r r r}
\toprule
& & \multicolumn{4}{c}{Ball-tree ordering} & \multicolumn{4}{c}{Random ordering} \\
\cmidrule(lr){3-6}\cmidrule(lr){7-10}
Benchmark & $K$ & median & p90 & p99 & max & median & p90 & p99 & max \\
\midrule
Elasticity     &  32 & 0.006523 & 0.0475  & 0.3238  & 0.3998  & 0.2035 & 0.2173 & 0.2274 & 0.2312 \\
Plasticity     &  64 & 0.00435  & 0.01515 & 0.1136  & 0.1675  & 0.1728 & 0.1909 & 0.1991 & 0.2045 \\
Airfoil        & 106 & 0.03324  & 133.3   & 435.9   & 450.6   & 77.3   & 118.3  & 154.3  & 155.2  \\
Pipe           &  64 & 0.04324  & 0.1611  & 6.678   & 17.65   & 8.937  & 9.383  & 10.18  & 10.6   \\
Navier--Stokes &  32 & 0.006677 & 0.006677& 0.006677& 0.006677& 0.171  & 0.1829 & 0.1866 & 0.1868 \\
Darcy          & 300 & 0.001223 & 0.003719& 0.05023 & 0.1706  & 0.1672 & 0.1728 & 0.1759 & 0.1905 \\
\bottomrule
\end{tabular}
\end{table}

\subsection{Choosing the Number of Patches K}
\label{app:chooseK}
Beyond efficiency considerations, the optimal number of patches $K$ depends on both the input resolution ($N$) and the nature of the underlying physical interactions.
Problems dominated by local interactions (e.g., solid/material settings such as Elasticity and Plasticity) typically require less global coupling, whereas fluid dynamics settings (e.g., Airfoil, Pipe, Navier--Stokes, Darcy) often benefit from stronger long-range communication due to advective transport and global pressure coupling.
In MSPT, $K$ controls a trade-off between (i) local context per patch, $L=N/K$, and (ii) global communication capacity via pooled tokens, proportional to $KQ$ supernodes.
At the extremes, $K\!\to\!N$ (so $L\!\to\!1$) approaches a global-attention regime similar in spirit to a full Transformer baseline, while $K=1$ reduces to a single patch with a pooled global token (analogous to a [CLS]-style global summary).

\subsection{Geometric Descriptors}
\label{app:geom_descriptors}

Following Transolver~\cite{wu2024Transolver}, we form input tokens by concatenating positional or geometric descriptors with the available physical-state features.
We use either raw coordinates or a distance-to-reference-grid descriptor
$g_i=\texttt{get\_grid}(\mathbf{pos})\in\mathbb{R}^{\texttt{ref}^d}$ (with $d\in\{2,3\}$ depending on whether the domain is 2D or 3D).
Across all experiments, we use the same geometric descriptors and preprocessing as Transolver. Additionally, we apply Rotary Positional Embeddings (RoPE) to the query and key projections within each patch prior to the dot-product attention~\cite{su2023roformerenhancedtransformerrotary}.

\subsection{Benchmarks}
The model is evaluated on eight benchmarks (see Table~\ref{tab:datasets}) covering three types of partial differential equations (PDEs): \textbf{Solid material} (Elasticity and Plasticity) \cite{dym1973solid}, \textbf{Navier-Stokes equations for fluid} \cite{mclean2012continuum} (Airfoil, Pipe, Navier-Stokes, Shape-Net Car, AhmedML), and \textbf{Darcy flow}\cite{hubbert1956darcy}. The specific details of each benchmark are provided below.

\vspace{-5pt}
\paragraph{Elasticity} This benchmark estimates internal stress in elastic materials from their structure, discretized into 972 points \cite{Li2022FourierNO}. Each input is a $972\times 2$ tensor for 2D positions, and the output is a $972\times 1$ stress tensor. There are 1000 training samples with various structures and 200 test samples.

\vspace{-5pt}
\paragraph{Plasticity} This benchmark predicts the future deformation of plastic materials subjected to impact from an arbitrarily shaped die \cite{Li2022FourierNO}. For each case, the input is the die shape, discretized into a structured mesh and represented as a $101\times 31$ tensor. The output is the deformation at each mesh point over 20 future time steps, recorded as a $20\times 101\times 31\times 4$ tensor, capturing deformation in four directions. The experimental protocol utilizes 900 training samples with diverse die shapes and 80 test samples.

\vspace{-5pt}
\paragraph{Airfoil} This task estimates the Mach number based on the airfoil shape, with the input discretized into a structured mesh of size $221\times 51$ and the output representing the Mach number at each mesh point \cite{Li2022FourierNO}. All shapes are derived from the NACA-0012 case provided by the National Advisory Committee for Aeronautics. The dataset comprises 1000 training samples with various airfoil designs and 200 samples for testing.

\vspace{-5pt}
\paragraph{Pipe} This benchmark estimates the horizontal fluid velocity based on the pipe structure \cite{Li2022FourierNO}. Each case discretizes the pipe into a structured mesh of size $129\times 129$. The input tensor, shaped $129\times 129\times 2$, encodes the positions of the discretized mesh points, while the output tensor, shaped $129\times 129\times 1$, represents the velocity at each point. The training set includes 1000 samples with varying pipe shapes, and 200 test samples are generated by modifying the pipe centerline.

\vspace{-5pt}
\paragraph{Navier-Stokes} This benchmark models incompressible and viscous flow on a unit torus, with constant fluid density and viscosity set to $10^{-5}$ \cite{FNO}. The fluid field is discretized into a $64\times 64$ regular grid. The task involves predicting the fluid state for the next 10 time steps based on observations from the previous 10 steps. The dataset consists of 1000 training samples with varying initial conditions and 200 samples for testing.

\vspace{-5pt}
\paragraph{Darcy} This benchmark models fluid flow through a porous medium \cite{FNO}. The process is initially discretized into a $421\times 421$ regular grid and subsequently downsampled to $85\times 85$ resolution for the main experiments, following \citet{wu2024Transolver}. The model input is the structure of the porous medium, and the output is the fluid pressure at each grid point. The dataset includes 1000 training samples and 200 test samples, each with distinct medium structures.

\vspace{-5pt}
\paragraph{Shape-Net Car} This benchmark addresses the estimation of drag coefficients for moving cars, a critical factor in automotive design. A total of 889 samples with diverse car shapes are generated to simulate driving at 72 km/h \cite{umetani2018learning}, using car models from the ShapeNet "car" category \cite{chang2015shapenet}. The simulation discretizes the space into an unstructured mesh with 32,186 points, capturing both the surrounding air and surface pressure. Following the experimental setup in 3D-GeoCA \cite{anonymous2023geometryguided}, 789 samples are used for training and 100 for testing. Each input mesh is preprocessed to include mesh point positions, signed distance functions, and normal vectors. The model predicts velocity and pressure at each point, enabling subsequent calculation of the drag coefficient from the estimated physical fields.

\vspace{-5pt}
\paragraph{AhmedML} AhmedML~\cite{ashton2024ahmed} provides 500 parameterized geometries simulated with a hybrid RANS--LES solver (OpenFOAM v2212) for 80 convective time units on $\sim\!20$\,M-cell unstructured meshes. Each case includes the body surface mesh, full volumetric fields (velocity and pressure), boundary slices, geometry parameters, and time-averaged forces.

\subsection{Metrics}

Since our experiment consists of standard benchmarks and practical design tasks, we also include several design-oriented metrics in addition to the relative $L_2$ for physics fields.

\vspace{-5pt}
\paragraph{Relative $L_2$ for physics fields} Given the physics field $\mathbf{u}$ and the model-predicted field $\widehat{\mathbf{u}}$, the relative $L_2$ of the model prediction is calculated as follows:
\begin{equation}
	\begin{split}
\operatorname{Relative\ L_2}=\frac{\|\mathbf{u}-\widehat{\mathbf{u}}\|}{\|\mathbf{u}\|}.
    \end{split}
\end{equation}
\vspace{-5pt}
\paragraph{Relative $L_2$ for drag coefficient}
For ShapeNet-Car, the drag coefficient is calculated based on the estimated physics fields, following the same implementation as \citet{wu2024Transolver}. For a unit-density fluid, the coefficient is defined as
\begin{equation}
C=\frac{2}{v^2A}\left(\int_{\partial\Omega}p(\boldsymbol{\xi})\left(\widehat{n}(\boldsymbol{\xi})\cdot\widehat{i}(\boldsymbol{\xi})\right)\mathrm{d}\boldsymbol{\xi} +\int_{\partial\Omega}\tau(\boldsymbol{\xi})\cdot\widehat{i}(\boldsymbol{\xi})\mathrm{d}\boldsymbol{\xi}\right),
\end{equation}
where $v$ is the speed of the inlet flow, $A$ is the reference area, $\partial\Omega$ is the object surface, $p$ denotes the pressure, $\widehat{n}$ is the outward unit normal vector of the surface, $\widehat{i}$ is the direction of the inlet flow, and $\tau$ denotes wall shear stress on the surface. $\tau$ can be calculated from the air velocity near the surface \cite{mccormick1994aerodynamics}, and is usually much smaller than the pressure term. For ShapeNet-Car, $\widehat{i}$ is set as $(-1,0,0)$ and $A$ is the area of the smallest rectangle enclosing the front of the car. The relative $L_2$ is defined between the ground-truth drag coefficient and the coefficient calculated from the predicted velocity and pressure fields.

\vspace{-5pt}
\paragraph{Spearman’s rank correlation for drag coefficient}
Given $M$ samples in the test set with ground-truth drag coefficients $T=\{T^{1},\cdots, T^{M}\}$ and model-predicted coefficients $\widehat{T}=\{\widehat{T}^{1},\cdots,\widehat{T}^{M}\}$, the Spearman correlation coefficient is defined as the Pearson correlation coefficient between the rank variables:
\begin{equation}
\rho=\frac{\operatorname{cov}\left(R(T),R(\widehat{T})\right)}{\sigma_{R(T)}\sigma_{R(\widehat{T})}},
\end{equation}
where $R$ is the ranking function, $\operatorname{cov}$ denotes the covariance, and $\sigma$ represents the standard deviation of the rank variables. This metric quantifies how well the model preserves the ordering of designs by drag, which is directly relevant for design optimization: higher $\rho$ indicates that it is easier to identify good designs from the model’s predictions \cite{spearman1904association}.

\subsection{Implementation Details}
\label{app:impl_details}

For the Standard PDE Benchmarks and ShapeNet-Car, we follow the training procedure outlined by \citet{wu2024Transolver}, using the \textit{Neural Solver Library}\footnote{\href{https://github.com/thuml/Neural-Solver-Library}{https://github.com/thuml/Neural-Solver-Library}}; MSPT does not restrict the training objective and can incorporate additional physics-informed regularizers (e.g., PDE residual, divergence, or boundary-condition penalties), although we report results with the standard benchmark objectives for fair comparison. For AhmedML, we adopt a training protocol similar to AB-UPT~\cite{alkin2025abuptscalingneuralcfd}, using 400 training samples, 50 validation samples, and 50 testing samples selected at random. To fit within a single A100 GPU (40 GB), each AhmedML mesh is partitioned into non-overlapping sub-meshes of 320k points (160k for surface and 160k for volume), enabling multiple blocks to be processed concurrently.
For models trained with the LION optimizer~\citep{lion}, we use a peak learning rate of $5\times10^{-5}$, weight decay of $5\times10^{-2}$, and a linear warmup over the first 5\% of training, followed by cosine decay to a final learning rate of $1\times10^{-6}$, as detailed in AB-UPT~\citep{alkin2025abuptscalingneuralcfd}. Models not trained with LION use the optimizer, and learning rate configurations implemented in the \textit{Neural Solver Library}. The complete training and model configurations for all benchmarks are summarized in Table~\ref{tab:training_model_detail}.

\begin{table}[H]
    \centering
    \caption{Training and model configurations of \METHOD. Here $\mathcal{L}_{\mathrm{v}}$ and $\mathcal{L}_{\mathrm{s}}$ represent the loss on volume and surface fields respectively. As for Darcy, we adopt an additional spatial gradient regularization term $\mathcal{L}_{\mathrm{g}}$ following ONO \cite{anonymous2023improved}.}
    \label{tab:training_model_detail}
    \vskip 0.1in
    \centering
    \begin{small}
    \begin{sc}
        \renewcommand{\multirowsetup}{\centering}
        \setlength{\tabcolsep}{2.2pt}
        \resizebox{\textwidth}{!}{
            \begin{tabular}{l|ccccc|cccccc}
                \toprule
                \multirow{3}{*}{Benchmarks} & \multicolumn{5}{c}{Training Configuration (Shared in all baselines)} & \multicolumn{6}{c}{Model Configuration} \\
                \cmidrule(lr){2-6}\cmidrule(lr){7-12} & Loss & Epochs & Initial LR & Optimizer & Batch Size & Layers $L$ & Heads & Channels $C$ & patches $K$ & Supernodes $Q$ & Pooling \\
                \midrule
                Elasticity & & \multirow{6}{*}{500} & $10^{-6}$ & LION~\cite{lion} & 1 & \multirow{6}{*}{8} & \multirow{6}{*}{8} & 128 & 32
                \\ Plasticity & & & \multirow{4}{*}{$10^{-3}$} & \multirow{4}{*}{AdamW~\cite{loshchilov2018decoupled}} & 8 & & & 128 & 64 \\
                Airfoil & Relative & & & & 4 & & & 128 & 106
                \\ Pipe & L2 & & & & 4 & & & 128 & 64 & 1 & Mean
                \\ Navier–Stokes & & & & & 2 & & & 256 & 32 \\
                Darcy & $\mathcal{L}_{\mathrm{rL2}}+0.1\mathcal{L}_{\mathrm{g}}$ & & $10^{-6}$ & LION~\cite{lion} & 1 & & & 128 & 300 \\
                \midrule
                Shape-Net Car & \multirow{2}{*}{$\mathcal{L}_{\mathrm{v}}+0.5\mathcal{L}_{\mathrm{s}}$} & 200 & \multirow{2}{*}{$10^{-6}$} & \multirow{2}{*}{LION~\cite{lion}} & \multirow{2}{*}{1} & \multirow{2}{*}{12} & \multirow{2}{*}{8} & \multirow{2}{*}{256} & 170 & \multirow{2}{*}{1} & \multirow{2}{*}{Mean} \\
                Ahmed ML & & 100 & & & & & & & 400
                \\
                \bottomrule
            \end{tabular}
        }
    \end{sc}
    \end{small}
    \vspace{-5pt}
\end{table}

\subsection{Efficiency Comparison with Baselines}
\label{app:efficiency_table}

\begin{table}[ht]
  \vspace{-5pt}
  \caption{Model efficiency comparison in Elasticity (Relative $L_2$) and Shape-Net Car ($\rho_{D}$), where we select Transolver and four other Transformer-based baseline methods that can be applied to unstructured meshes. Baselines are taken from \citet{wu2024Transolver}. Efficiency is evaluated on inputs of different meshes during training. Running time is measured by the time to complete one epoch, which contains $10^3$ iterations. ``/'' indicates the baseline will fail in this benchmark. For a like-for-like comparison, we set MSPT to $K=32$ patches to match Transolver's $S=32$ slices; unlike Transolver whose runtime increases with larger $S$, MSPT can reduce runtime by increasing $K$ (i.e., using smaller patches).}
  \label{tab:all_efficiency}
  \vskip 0.05in
  \centering
  \begin{threeparttable}
  \begin{small}
  \renewcommand{\multirowsetup}{\centering}
  \setlength{\tabcolsep}{7pt}
  \begin{tabular}{c|ccc|c|c}
    \toprule
    \multirow{2}{*}{Model} & Input Mesh Size & GPU Memory & Running Time & Elasticity  & Shape-Net Car \\
    \cmidrule{5-6}
     & $N$ & (GB) & (s~/~epoch) & (972 mesh points) & (32,186 mesh points) \\
    \toprule
    \multirow{8}{*}{\textbf{MSPT (Ours)}} & 1024  & 0.17 & 46.6095 & \multirow{8}{*}{0.0048} & \multirow{8}{*}{0.9941} \\
    & 2048  & 0.22 & 46.8275 &  &  \\
    & 4096  & 0.33 & 46.6800 &  &  \\
    & 8192  & 0.53 & 46.4544 &  &  \\
    & 16384 & 0.89 & 47.1659 &  &  \\
    & 32768 & 1.65 & 65.0738 &  &  \\
    & 100000 & 4.68 & 192.5923 &  &  \\
    & 200000 & 9.13 & 365.9022 &  &  \\
    & 500000 & 22.49 & 797.7500 &  &  \\
    \midrule
    \multirow{8}{*}{Transolver~\citep{wu2024Transolver}} & 1024  & 0.13 & 36.5752 & \multirow{8}{*}{0.0064} & \multirow{8}{*}{0.9935} \\
    & 2048  & 0.20 & 36.9748 &  &  \\
    & 4096  & 0.34 & 37.3330 &  &  \\
    & 8192  & 0.63 & 36.5046 &  &  \\
    & 16384 & 1.21 & 37.3509 &  &  \\
    & 32768 & 2.37 & 41.6690 &  &  \\
    & 100000 & 7.05 & 104.3364 &  &  \\
    & 200000 & 14.06 & 199.3195 &  &  \\
    & 500000 & 34.94 & 547.3848 &  &  \\
    \midrule
    \midrule
    & 1024  & 0.85 & 54.282  & \multirow{5}{*}{0.0086} & \multirow{5}{*}{0.9833} \\
    & 2048  & 1.07 & 55.939  &  &  \\
    GNOT~\citep{hao2023gnot} & 4096  & 1.47 & 60.857  &  &  \\
    & 8192  & 2.33 & 67.170  &  &  \\
    & 16384 & 4.23 & 112.552 &  &  \\
    & 32768 & 7.46 & 209.923 &  &  \\
    \midrule
    & 1024  & 1.47 & 69.759  & \multirow{5}{*}{0.0118} & \multirow{5}{*}{/} \\
    & 2048  & 1.75 & 76.245  &  &  \\
    ONO~\citep{anonymous2023improved} & 4096  & 2.30 & 100.134 &  &  \\
    & 8192  & 3.47 & 149.598 &  &  \\
    & 16384 & 5.64 & 255.339 &  &  \\
    & 32768 & 10.09 & 462.459 &  &  \\
    \midrule
    & 1024  & 0.63 & 28.147  & \multirow{5}{*}{0.0183} & \multirow{5}{*}{/} \\
    & 2048  & 0.69 & 30.983  &  &  \\
    OFormer~\citep{li2023transformer} & 4096  & 0.80 & 31.113  &  &  \\
    & 8192  & 1.02 & 47.904  &  &  \\
    & 16384 & 1.67 & 91.671  &  &  \\
    & 32768 & 2.44 & 182.205 &  &  \\
    \midrule
    & 1024  & 0.62 & 26.507  & \multirow{5}{*}{0.0240} & \multirow{5}{*}{0.9764} \\
    & 2048  & 0.66 & 26.503  &  &  \\
    Galerkin Transformer~\citep{Cao2021ChooseAT} & 4096  & 0.74 & 27.481  &  &  \\
    & 8192  & 0.91 & 37.098  &  &  \\
    & 16384 & 1.45 & 67.524  &  &  \\
    & 32768 & 2.05 & 129.872 &  &  \\
    \bottomrule
  \end{tabular}
  \end{small}
  \end{threeparttable}
  \vspace{-5pt}
\end{table}

\newpage
\section{Additional Visualizations}
\label{app:app_pde_benchmark}

In this section, we provide additional results comparing the performance of \METHOD\ and Transolver on a set of benchmark PDE problems.

\begin{figure}[ht]
    \centering
    \newcommand{\imgheightA}{3.5cm}
    \newcommand{\rowtitleheight}{1\baselineskip}

    \begin{minipage}[t]{\linewidth}
        \centering
        \parbox[t][\rowtitleheight][t]{\linewidth}{\centering \small \textbf{Airfoil}}\\[3pt]
        \begin{minipage}[t]{0.32\linewidth}
            \centering
            \includegraphics[height=\imgheightA]{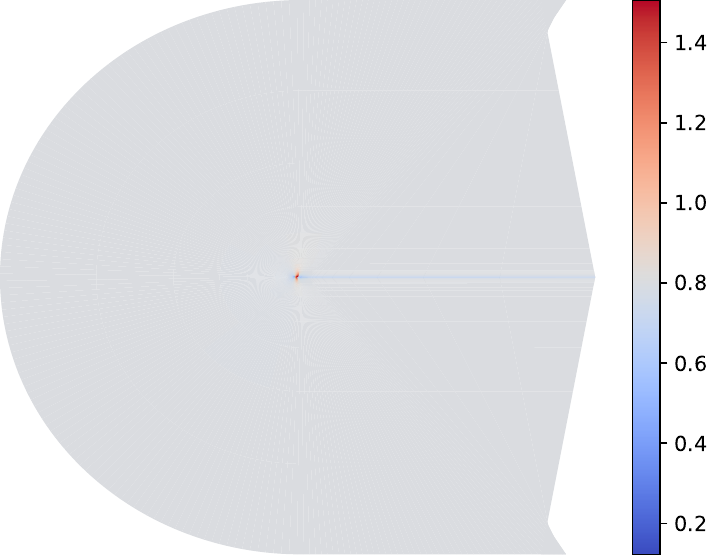}\\[2pt]
            \hspace{-0.5em}\small Ground-truth
        \end{minipage}\hfill
        \begin{minipage}[t]{0.32\linewidth}
            \centering
            \includegraphics[height=\imgheightA]{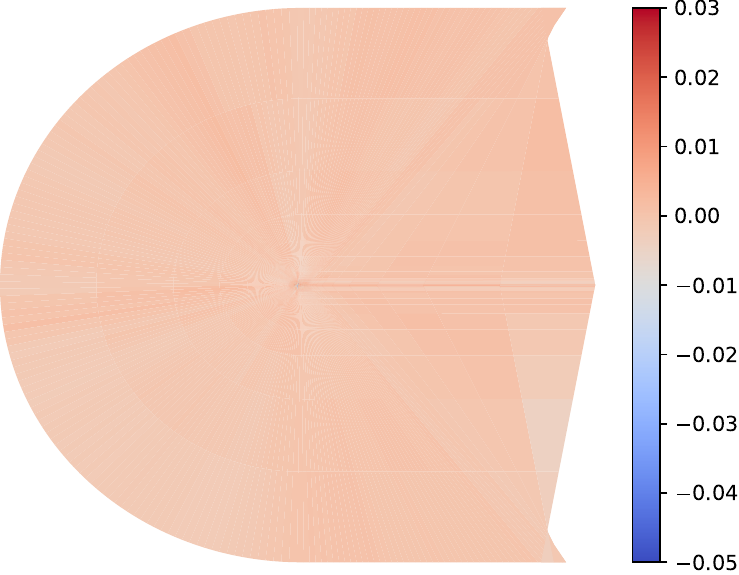}\\[2pt]
            \hspace{-0.5em}\small \METHOD{} (Ours)
        \end{minipage}\hfill
        \begin{minipage}[t]{0.32\linewidth}
            \centering
            \includegraphics[height=\imgheightA]{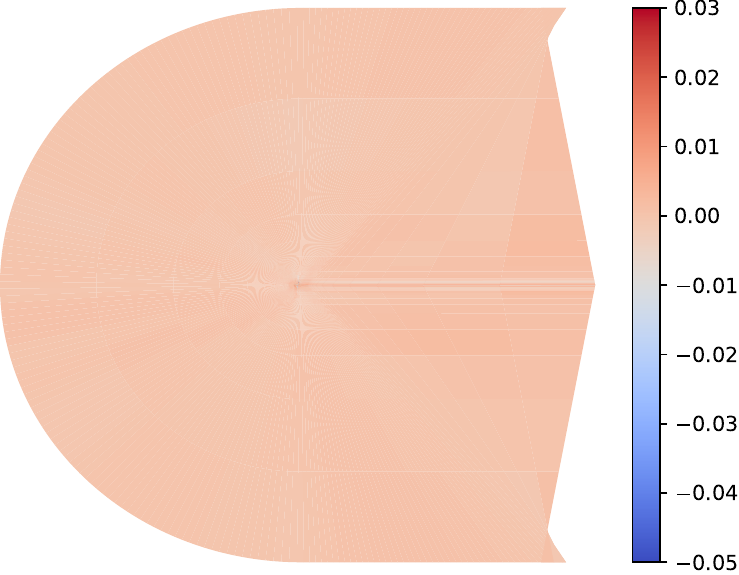}\\[2pt]
            \hspace{-0.5em}\small Transolver
        \end{minipage}
    \end{minipage}

    \vspace{0.9em}

    \begin{minipage}[t]{\linewidth}
        \centering
        \parbox[t][\rowtitleheight][t]{\linewidth}{\centering \small \textbf{Elasticity}}\\[3pt]
        \begin{minipage}[t]{0.32\linewidth}
            \centering
            \includegraphics[height=\imgheightA]{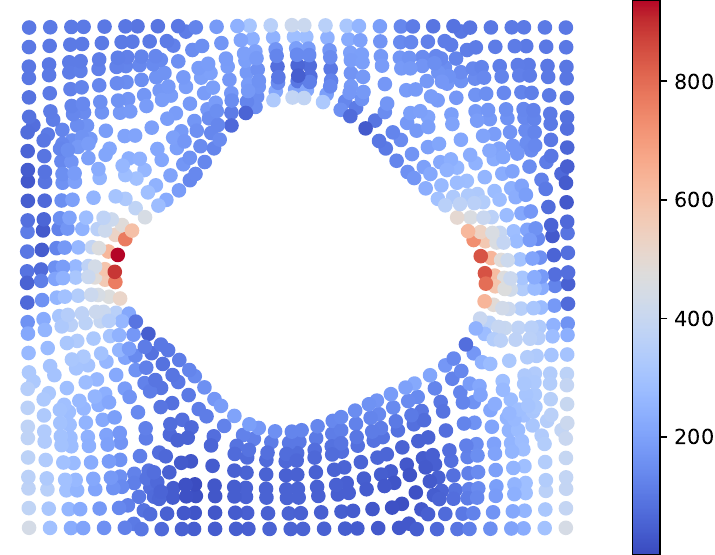}\\[2pt]
            \hspace{-0.5em}\small Ground-truth
        \end{minipage}\hfill
        \begin{minipage}[t]{0.32\linewidth}
            \centering
            \includegraphics[height=\imgheightA]{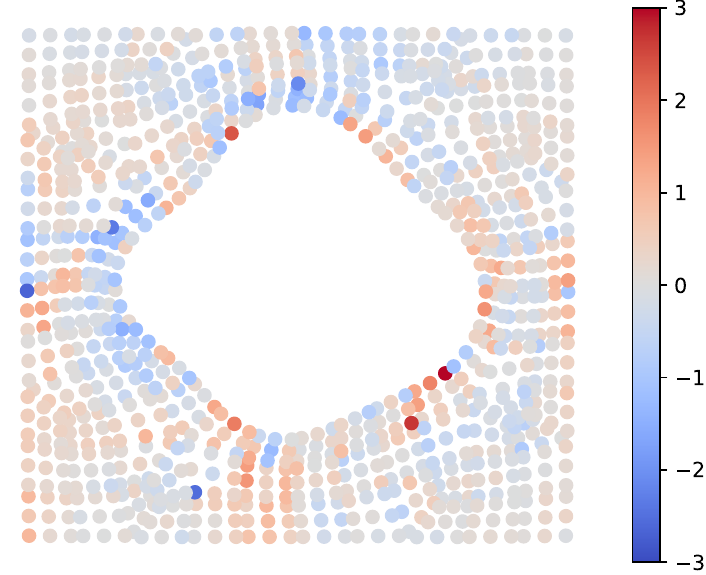}\\[2pt]
            \hspace{-0.5em}\small \METHOD{} (Ours)
        \end{minipage}\hfill
        \begin{minipage}[t]{0.32\linewidth}
            \centering
            \includegraphics[height=\imgheightA]{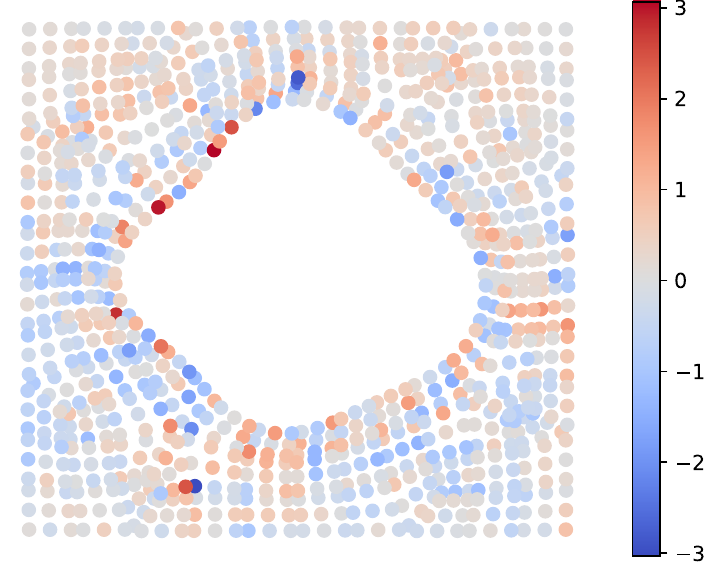}\\[2pt]
            \hspace{-0.5em}\small Transolver
        \end{minipage}
    \end{minipage}

    \vspace{0.9em}

    \begin{minipage}[t]{\linewidth}
        \centering
        \parbox[t][\rowtitleheight][t]{\linewidth}{\centering \small \textbf{Darcy}}\\[3pt]
        \begin{minipage}[t]{0.32\linewidth}
            \centering
            \includegraphics[height=\imgheightA]{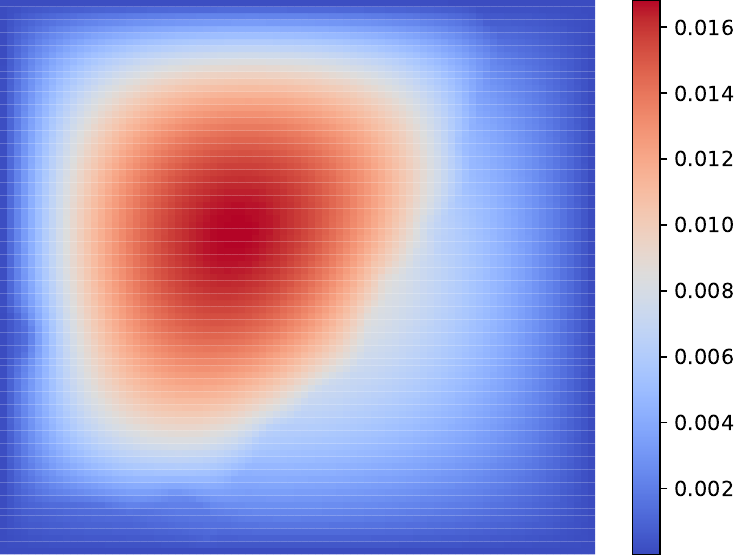}\\[2pt]
            \hspace{-0.5em}\small Ground-truth
        \end{minipage}\hfill
        \begin{minipage}[t]{0.32\linewidth}
            \centering
            \includegraphics[height=\imgheightA]{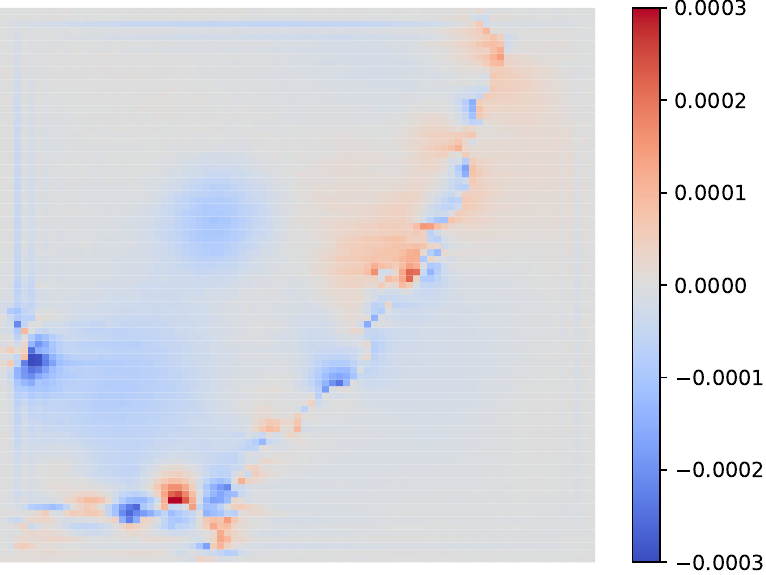}\\[2pt]
            \hspace{-0.5em}\small \METHOD{} (Ours)
        \end{minipage}\hfill
        \begin{minipage}[t]{0.32\linewidth}
            \centering
            \includegraphics[height=\imgheightA]{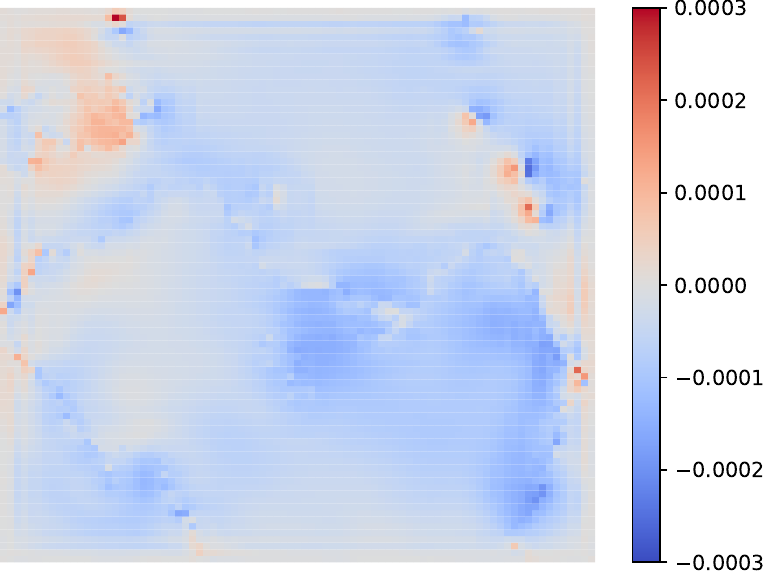}\\[2pt]
            \hspace{-0.5em}\small Transolver
        \end{minipage}
    \end{minipage}

    \caption{
    Visual comparison of per-point relative error maps across standard PDE benchmarks.
    Each row corresponds to one dataset (Airfoil, Elasticity, Darcy), and columns show the
    ground-truth field, \METHOD{}, and Transolver.
    }
    \label{fig:pde_benchmarks_standard}
\end{figure}

\newpage
\section{Reproducibility Results}
\label{app:app_reproducibility}

\subsection{Transolver++}

Transolver++~\citep{luo2025transolverpp} is a parallelizable and more efficient successor to Transolver~\citep{wu2024Transolver}. 
We trained Transolver++ using the official implementation\footnote{\href{https://github.com/thuml/Transolver_plus/tree/main}{https://github.com/thuml/Transolver\_plus}} and adapted it to all geometries following the Transolver setup, with only minimal adjustments for structured meshes. Training was carried out using the Neural-Solver-Library framework\footnote{\href{https://github.com/thuml/Neural-Solver-Library}{https://github.com/thuml/Neural-Solver-Library}}, matching the experimental settings reported in \citet{luo2025transolverpp}.

As shown in Tables~\ref{tab:repro_pdebench} and~\ref{tab:repro_shapenetcar}, Transolver++ (\textit{repr.}) performs worse than both Transolver and \METHOD{}. Similar reproduction differences were also reported by AB-UPT~\citep{alkin2025abuptscalingneuralcfd}. To avoid over-interpreting discrepancies, we do not include Transolver++ in the main benchmark comparisons and instead report its reproduced results here for completeness.

\begin{table}[h]
    \caption{\textbf{Reproducibility comparison on standard PDE benchmarks.} Relative $L_2$ errors are reported. 
    ``Transolver++ (repr.)'' refers to our reproduction using the official implementation. 
    ``/'' indicates inapplicability. 
    All values are shown in units of $\times 10^{-2}$.}
    \label{tab:repro_pdebench}
    \centering
    \setlength{\tabcolsep}{3pt}
    \begin{tabular}{@{}l@{}cccccc@{}}
        \toprule
        \multirow{4}{*}{Model} & Point Cloud & \multicolumn{3}{c}{Structured Mesh} & \multicolumn{2}{c}{Regular Grid} \\
        \cmidrule(lr){2-2}\cmidrule(lr){3-5}\cmidrule(lr){6-7}
        & \multirow{2}{*}{Elasticity} & \multirow{2}{*}{Plasticity} & \multirow{2}{*}{Airfoil} & \multirow{2}{*}{Pipe} & Navier & \multirow{2}{*}{Darcy} \\
        & & & & & Stokes & \\
        \midrule
        Transolver~\citep{wu2024Transolver} & 0.64 & 0.12 & 0.53 & 0.33 & 9.00 & 0.57 \\
        Transolver++~\citep{luo2025transolverpp} & 0.52 & 0.11 & 0.48 & 0.27 & 7.19 & 0.49 \\
        Transolver++ (\textit{repr.})~\citep{luo2025transolverpp} & 1.54 & 0.54 & 0.75 & 0.55 & 12.30 & 0.90 \\
        \midrule
        \textbf{\METHOD\ (Ours)} & 0.48 & 0.10 & 0.51 & 0.31 & 6.32 & 0.63 \\
        \bottomrule
    \end{tabular}
\end{table}

\begin{table}[ht]
    \caption{\textbf{Reproducibility comparison on ShapeNet-Car.} Relative $L_2$ errors for volume and surface fields, and drag coefficient ($C_D$); Spearman $\rho_D$ for ranking quality. 
    ``Transolver++ (repr.)'' refers to our reproduction using the official implementation. 
    All models are trained under identical experimental conditions. 
    Values are shown in units of $\times 10^{-2}$.}
    \label{tab:repro_shapenetcar}
    \centering
    \setlength{\tabcolsep}{4pt}
    \begin{tabular}{@{}lcccc@{}}
        \toprule
        \multirow{2}{*}{\textbf{Model}} & \multicolumn{4}{c}{\textbf{ShapeNet-Car}} \\
        \cmidrule(lr){2-5}
        & Volume $\downarrow$ & Surf $\downarrow$ & $(C_D)$ $\downarrow$ & $(\rho_D)$ $\uparrow$ \\
        \midrule
        Transolver~\citep{wu2024Transolver} & 2.07 & 7.45 & 1.03 & 99.35 \\
        Transolver++ (\textit{repr.})~\citep{luo2025transolverpp} & 2.37 & 8.37 & 1.41 & 99.24 \\
        \midrule
        \textbf{\METHOD\ (Ours)} & 1.89 & 7.41 & 0.98 & 99.41 \\
        \bottomrule
    \end{tabular}
\end{table}

\subsection{AB-UPT}

Since the official AB-UPT~\citep{alkin2025abuptscalingneuralcfd} training pipeline is not publicly available, we reproduced its results using the \textit{Neural-Solver-Library} (following the same training procedure as~\citet{wu2024Transolver}) for the ShapeNet-Car dataset, and our own training pipeline for AhmedML, following the model configurations and hyperparameters (e.g., optimizer, learning rate, batch size) reported in their paper. The reproduced results differ from the originally published values, which is consistent with the authors’ note that their framework applies shared modifications across all models and that deviations can arise from training randomness and implementation details. In our experiments, similar discrepancies also appear when retraining other baselines under a unified pipeline, suggesting that part of the gap is due to differences in training frameworks rather than the architectures themselves. Accordingly, in the main text we report the original AB-UPT numbers for consistency with their publication, and provide our reproduced results and analysis here for completeness.

\end{document}